\documentclass[10pt,twocolumn,letterpaper]{article}

\usepackage{iccv}          
\usepackage{algorithm}
\usepackage{algpseudocode}
\usepackage{multirow}
\usepackage{tikz} 
\usepackage{array}
\usepackage{float}
\usepackage{placeins}
\usepackage{fontawesome5}        
\usepackage{graphicx}

\definecolor{iccvblue}{rgb}{0.21,0.49,0.74}
\usepackage[pagebackref,breaklinks,colorlinks,allcolors=iccvblue]{hyperref}

\def\paperID{7344} 
\def\confName{ICCV}
\def\confYear{2025}

\title{MSQ: Memory-Efficient Bit Sparsification Quantization}

\author{
Seokho Han\textsuperscript{1*} \quad
Seoyeon Yoon\textsuperscript{1*} \quad
Jinhee Kim\textsuperscript{1} \quad
Dongwei Wang\textsuperscript{2}\\
Kang Eun Jeon\textsuperscript{1\faEnvelope[regular]} \quad
Huanrui Yang\textsuperscript{2\faEnvelope[regular]} \quad
Jong Hwan Ko\textsuperscript{1\faEnvelope[regular]} \\
$^{1}$Department of Electrical and Computer Engineering, Sungkyunkwan University, Korea \\
$^{2}$Department of Electrical and Computer Engineering, University of Arizona, USA \\
{\tt\small \{beppa2396, syy000405, a2jinhee, kejeon, jhko\}@skku.edu} \\
{\tt\small \{dongweiw, huanruiyang\}@arizona.edu}
}

\newcommand{\mysubsection}[1]{\vspace{0.5em}\noindent\textbf{#1}}

\begin{document}
\maketitle
\renewcommand{\thefootnote}{\fnsymbol{footnote}}
\footnotetext[1]{: Equal contributions}
\renewcommand{\thefootnote}{\faEnvelope[regular]}
\footnotetext[2]{: Corresponding authors}
\begin{abstract}
As deep neural networks (DNNs) see increased deployment on mobile and edge devices, optimizing model efficiency has become crucial. Mixed-precision quantization is widely favored, as it offers a superior balance between efficiency and accuracy compared to uniform quantization. However, finding the optimal precision for each layer is challenging. Recent studies utilizing bit-level sparsity have shown promise, yet they often introduce substantial training complexity and high GPU memory requirements. In this paper, we propose Memory-Efficient Bit Sparsification Quantization (MSQ), a novel approach that addresses these limitations. 
MSQ applies a round-clamp quantizer to enable differentiable computation of the least significant bits (LSBs) from model weights. It further employs regularization to induce sparsity in these LSBs, enabling effective precision reduction without explicit bit-level parameter splitting. 
Additionally, MSQ incorporates Hessian information, allowing the simultaneous pruning of multiple LSBs to further enhance training efficiency. Experimental results show that MSQ achieves up to 8.00× reduction in trainable parameters and up to 86\% reduction in training time compared to previous bit-level quantization, while maintaining competitive accuracy and compression rates. This makes it a practical solution for training efficient DNNs on resource-constrained devices.
\end{abstract}

\section{Introduction}
\label{sec:intro}

With the increasing adoption of deep neural networks (DNNs) in mobile and edge devices~\cite{sandler2018mobilenetv2,hu2018squeeze}, optimizing their efficiency has become a major area of research due to their limited computational and memory resources~\cite{burrello2021dory}. Quantization - transforming model weights (and activations) from high-precision floating-point to low-precision fixed-point values~\cite{yang2019quantization, banner2019post,zhou2016dorefa,choi2018pact, zhang2018lq} - has been widely adopted to address this challenge. This transformation not only reduces the storage requirements of models but also enables the use of fixed-point arithmetic units, significantly lowering energy and area costs while improving computational speed~\cite{horowitz20141}.

However, quantization introduces noises in model weights and activations due to the discrepancy between the original floating-point and low-precision values, potentially degrading model performance. When every layer of a model is quantized to the same low precision, the noise accumulated from some sensitive layers can lead to substantial accuracy drops. Mixed-precision quantization, which assigns different precision levels to each layer, has demonstrated improvements in achieving higher accuracies with smaller model sizes~\cite{wu2019fbnet}.

Despite its advantages in performance and compression rate, mixed-precision quantization is challenging to implement due to the vast search space required to determine the optimal bit scheme. Previous approaches have utilized reinforcement learning-based searches~\cite{wang2019haq} or computed sensitivity statistics on pre-trained models~\cite{dong2019hawq,dong2020hawq} to assign higher precision to sensitive layers and lower precision to less critical ones. However, these search-based methods are resource-intensive and fail to capture sensitivity changes during quantization-aware finetuning.To address this, dynamic mixed-precision quantization adjustment via bit-level structural sparsity has gained attention~\cite{yang2021bsq,xiao2023csq}. These methods induce certain bits in the fixed-point weight representation to zeros, enabling simultaneous weight training and quantization scheme adjustment. However, previous bit-level training methods require instantiating a trainable variable for each individual bit, leading to extended training times and high GPU memory consumption.

This work aims to mitigate the memory and training cost challenges of the bit-level training to achieve mixed-precision quantization scheme more efficiently. We propose Memory-Efficient Bit Sparsification Quantization (MSQ), a method to induce bit-level sparsity without explicitly introducing bit-level trainable parameters. 
% MSQ employs a novel \textit{round-clamp quantizer} to allow simple least significant bits (LSBs) computation directly from the floating-point latent weight. $\ell_1$ regularization can then be applied on the computed LSBs to effectively introduce sparsity and, consequently, precision reductions. 
MSQ derives LSB sparsity and aims to prune it. To achieve this, it employs a novel \textit{round-clamp quantizer} provides a gradient direction  for LSB sparsification. Furthermore, $\ell_1$ regularization is applied to the computed LSB to effectively induce sparsity and, consequently, enable precision reduction.
Additionally, MSQ incorporates Hessian information to account for layer sensitivity, enabling faster bit pruning rate on insensitive layers for greater training efficiency.The experiment examines training efficiency and the accuracy-compression tradeoff. Training with fewer parameters ,due to the absence of bit-level splitting, achieves significant memory savings and reduces training time by up to 86\% compared to the traditional bit-level splitting approach, while experiments on ResNet and ViT models demonstrate a comparable accuracy-compression tradeoff.

Our contributions with MSQ are as follows:
\begin{itemize}
	\item Significantly reducing bit-level training and sparsification cost by mitigating explicit bit splitting.
	\item Introducing a novel round-clamp quantizer for effective LSB computation and sparsity-inducing regularization.
    \item Extending bit-level quantization to both heterogeneous CNNs (e.g., MobileNetV3) and complex architectures such as Vision Transformers (ViTs).

\end{itemize}

\section{Related Works}
\label{sec:relatedwork}
\mysubsection{Quantization and quantization-aware training.} Quantization techniques transform floating-point weight parameters into integer representations with reduced numerical precision. Though post-training quantization, which directly applies quantization to a pre-trained model, has undergone tremendous improvements~\cite{nagel2020up,liu2023noisyquant,xiao2023smoothquant,guo2022squant},
%(PTQ example) 
finetuning is still required under ultra-low precision scenarios to prevent substantial model performance degradation. Consequently, quantization-aware training (QAT) techniques have been investigated to finetune quantized models under low-precisions. Since quantized weights assume discrete values, traditional gradient-based optimization methods, designed for continuous parameter spaces, are not directly applicable to training quantized models~\cite{zhou2016dorefa,choi2018pact,yang2021bsq,dong2019hawq}. To address this challenge, algorithms such as DoReFa-Net~\cite{zhou2016dorefa} employ the straight-through estimator (STE) to allow continuous gradient accumulation in the floating-point format. 
The forward and backward computation of the quantization-aware training process can be expressed as:
\begin{equation}
    \mathbf{W_n} = \frac{1}{2^n - 1} \operatorname{Round}[(2^n - 1) \mathbf{W}],
\label{eq:ste}
\end{equation}
\begin{equation}
\frac{\partial \mathcal{L}}{\partial \mathbf{W}} = \frac{\partial \mathcal{L}}{\partial \mathbf{W_n}},
\label{eq:ste_grad}
\end{equation}
where a normalized floating-point weight $\mathbf{W}$ is kept throughout the entire training. Along the forward pass, the STE will quantize $\mathbf{W}$ to an \( n \)-bit fixed-point representation $\mathbf{W_n}$, which will be used to compute the model output and loss \( \mathcal{L} \). During the backward pass, the STE will directly pass the gradient w.r.t. $\mathbf{W_n}$ onto $\mathbf{W}$, which enables $\mathbf{W}$ to be updated with the standard gradient-based optimizer.

\mysubsection{Mixed-precision quantization}
is explored following the observation that different layers in the model contributes differently to the model loss~\cite{dong2019hawq,wang2019haq}. A central challenge in mixed-precision quantization research is designing an optimal set of bit schemes to balance performance and model size. Initial approaches relied on manual heuristics, such as setting higher precision for the first and last layers. Search-based methods, like HAQ~\cite{wang2019haq}, use reinforcement learning to determine the quantization scheme. This process can be costly, especially for deeper models with a large search space. Other methods attempt to measure each layer's sensitivity directly, using metrics such as Hessian eigenvalue or trace~\cite{dong2019hawq,dong2020hawq}. However, these approaches only account for the sensitivity of a fully trained, high-precision model and do not consider sensitivity changes that occur during quantization or updates in quantization-aware training.

\mysubsection{Bit-level quantization.}
\begin{figure}[t]
    \centering
    \includegraphics[width=1.0\columnwidth]{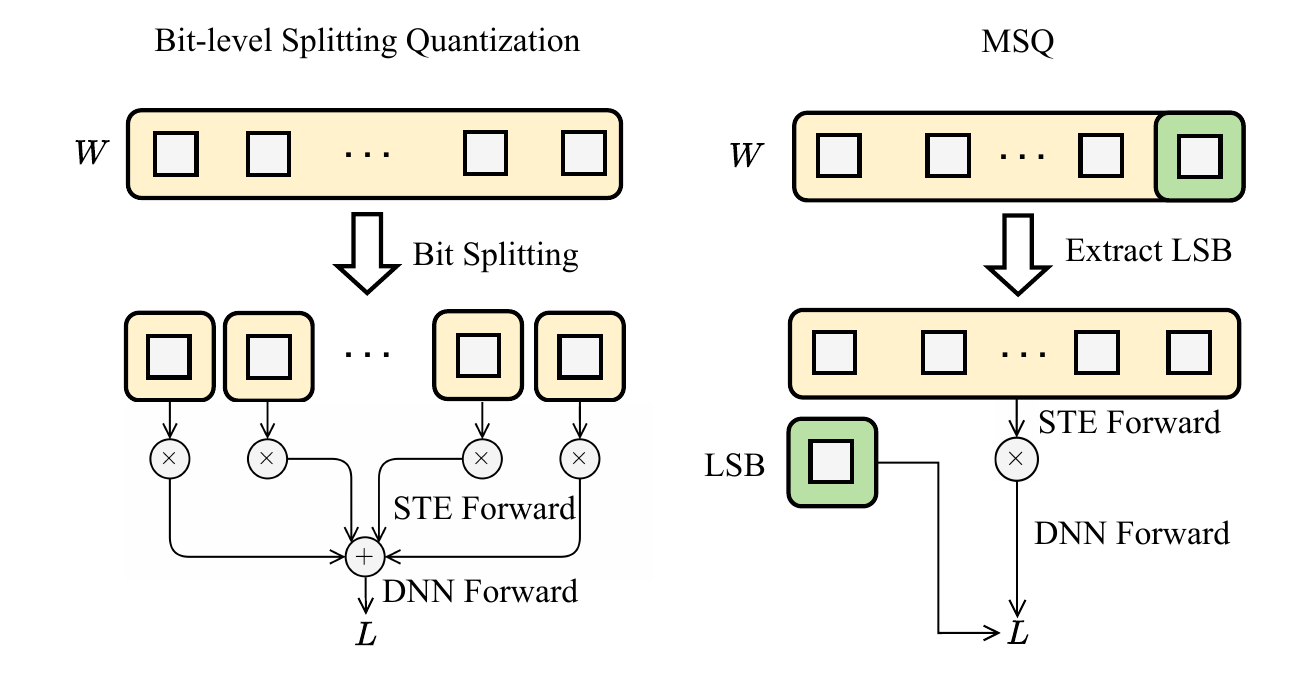}
    \caption{Work flow of bit-level quantization vs. MSQ.}
    \label{fig:RW_bit_level}
\end{figure}
%%%%%%%%%%%%%%%%%%%%%%%%%%%%%%%%%%%%%%%%%%%%%%%%%%%%%%%%%%%%%%%%%%%%%%%%%%%%%
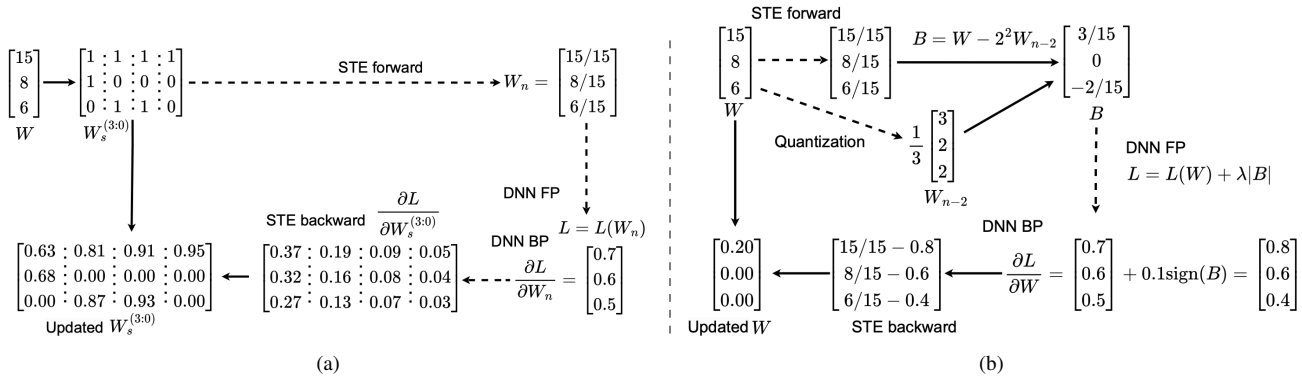
\begin{figure*}[t]
  \centering
  \begin{subfigure}{0.51\linewidth}
    \centering
    \includegraphics[width=1\linewidth]{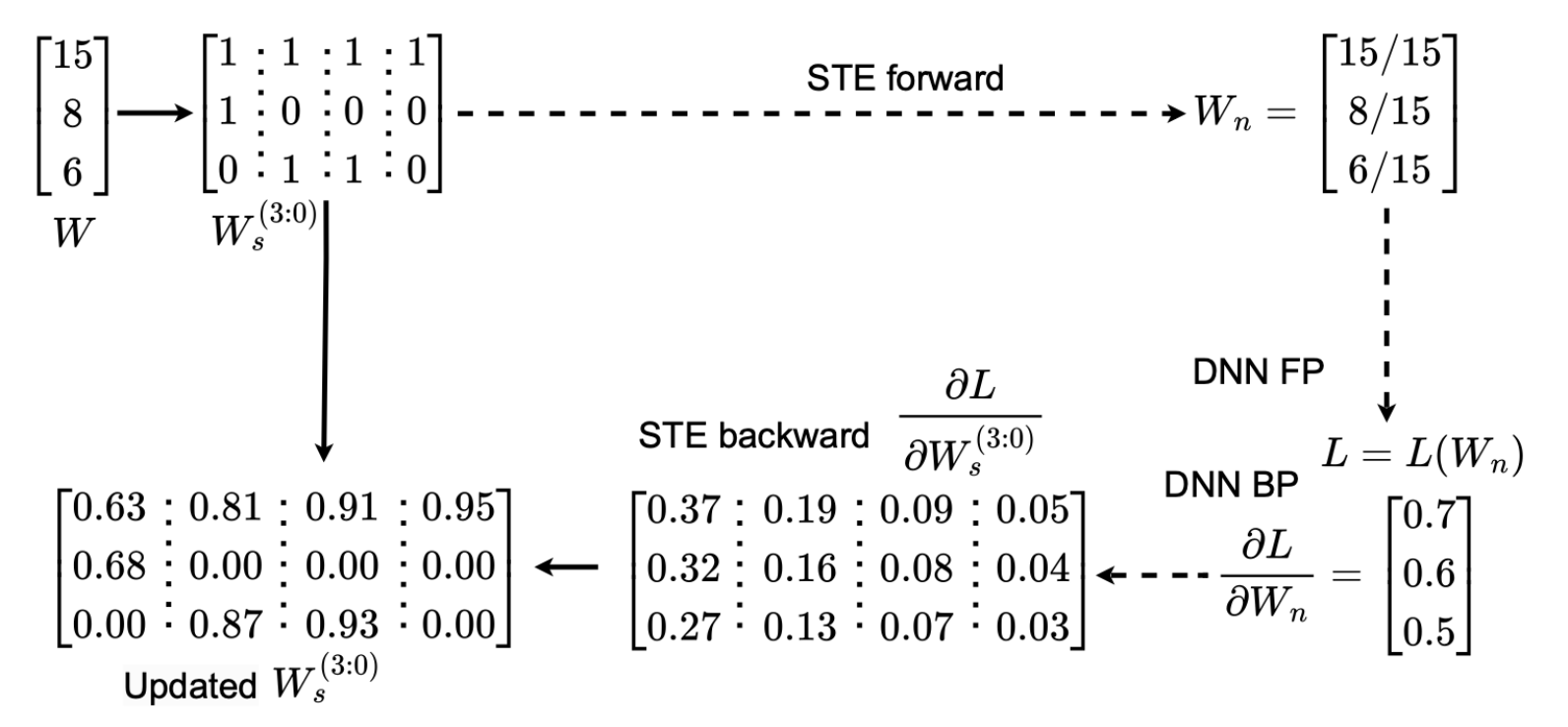}
    \caption{}
    \label{fig:bsq_train}
  \end{subfigure}
    \begin{minipage}{0.0001\linewidth}
    \centering
        \vspace{-50mm}
    \begin{tikzpicture}
      \draw[dashed, line width=0.1mm] (0,0) -- (0,4);
    \end{tikzpicture}
  \end{minipage}
  \begin{subfigure}{0.48\linewidth}
    \centering
    \includegraphics[width=1\linewidth]{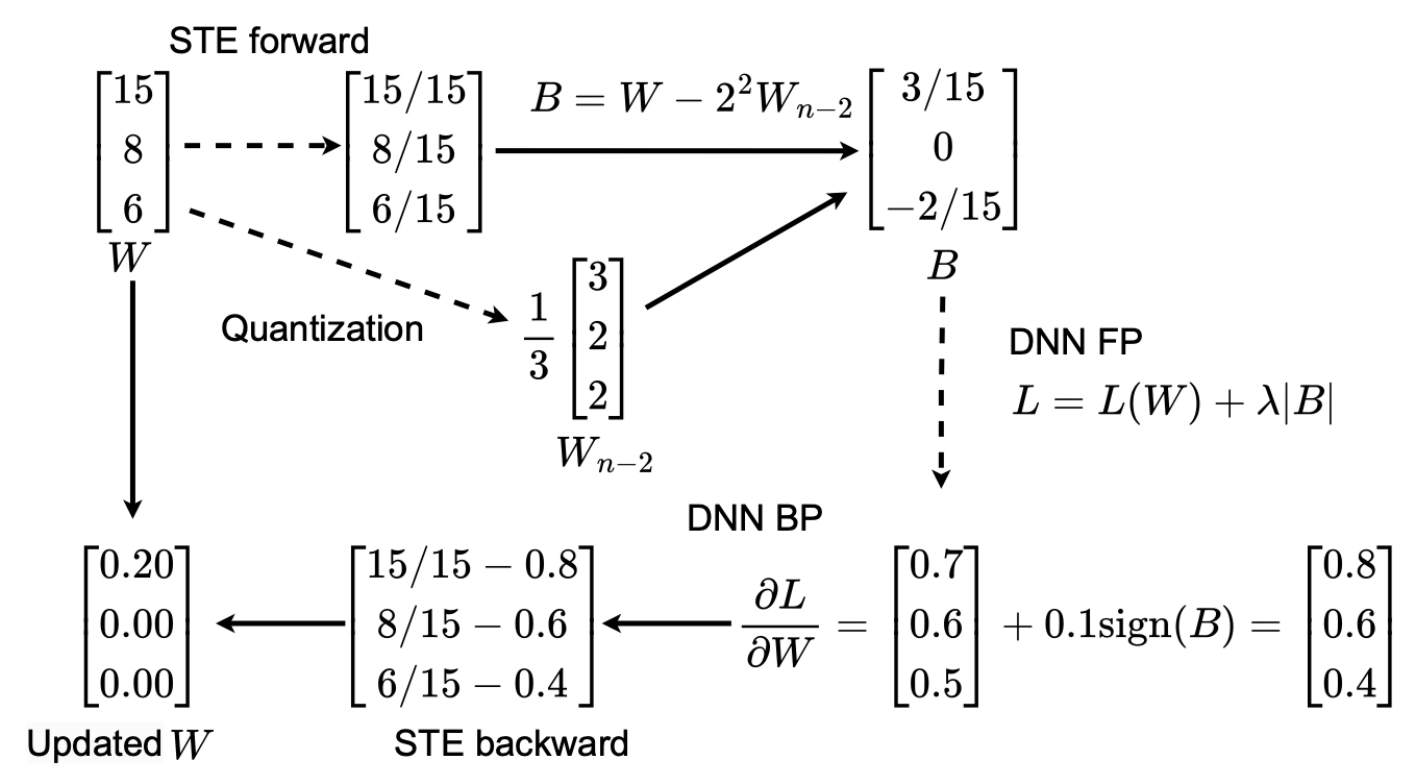}
    \caption{}
    \label{fig:msq_train}
  \end{subfigure}
 \vspace{-5mm}
  \caption{Comparison of bit-level splitting quantization and MSQ. (a) Training process of the bit-level model weight with STE.(b) Training process of the MSQ with LSB regularization.}
  \label{fig:train}
\end{figure*}
%%%%%%%%%%%%%%%%%%%%%%%%%%%%%%%%%%%%%%%%%%%%%%%%%%%%%%%%%%%%%%%%%%%%%%%%%%%%%
Bit-level sparsity quantization (BSQ)~\cite{yang2021bsq} was the first method to simultaneously apply a mixed-precision quantization scheme and train a quantized DNN model within a single training phase. In BSQ, each bit in the quantized model is treated as an independent trainable variable, where bit-level structural sparsity is induced in training to achieve mixed-precision quantization schemes. Although BSQ has shown strong empirical results, it requires straight-through gradient estimation for the bit variables in the rounded bit-level representation, which can reduce accuracy. Additionally, hard precision adjustments applied through bit pruning during training can compromise convergence stability. To mitigate this instability, Continuous Sparsification Quantization (CSQ) ~\cite{xiao2023csq} was introduced, which smooths both bit-level training and precision adjustments using continuous sparsification ~\cite{savarese2020winning}.

While bit-level quantization allows for simultaneous mixed-precision quantization scheme discovery and parameter training, it demands substantial resources, including increased time and GPU memory, due to the large number of trainable parameters required for bit-level treatment as shown in Fig.~\ref{fig:RW_bit_level}. In this work, we reduce these resource demands by eliminating the need for multiplied parameters and instead applying a direct LSB computation and regularization from original trainable parameters, resulting in lower training costs compared to prior bit-level quantization methods.

\section {Method}
\label{sec:method}

In this work, we propose MSQ, an efficient training algorithm to achieve bit-level sparsity for mixed-precision quantization scheme. MSQ mitigates the burden of having independent bit-level trainable variables as in BSQ~\cite{yang2021bsq} and CSQ~\cite{xiao2023csq} by deriving the sparsity of LSBs and prune them. To achieve this, we introduce a novel RoundClamp quantizer that supports bipartite bit-slicing for LSB computation and regularization, and regularization in \cref{sec:bbs} We then propose the Hessian-aware aggressive pruning technique to control the layer-wise bit reduction speed in \cref{sec:haap} to reach the targeted model compression ratio with less training epochs. Finally, the full procedure and details of the proposed MSQ algorithm is discussed in \cref{sec:ota}. The training process of MSQ is illustrated in \cref{fig:train}.

\subsection{Bipartite Bit Slicing}
\label{sec:bbs}
\noindent\textbf{RoundClamp quantizer design.} 
In previous work of BSQ and CSQ, we observe that most precision reduction happens when LSBs are removed. This comes from the fact that LSBs has the minimal impact on the quantized weight values, therefore are induced to zeros more easily by the sparsity-inducing regularizer.

\begin{figure}[tb]
    \centering
    \begin{subfigure}{1\linewidth}
         \includegraphics[width=\linewidth]{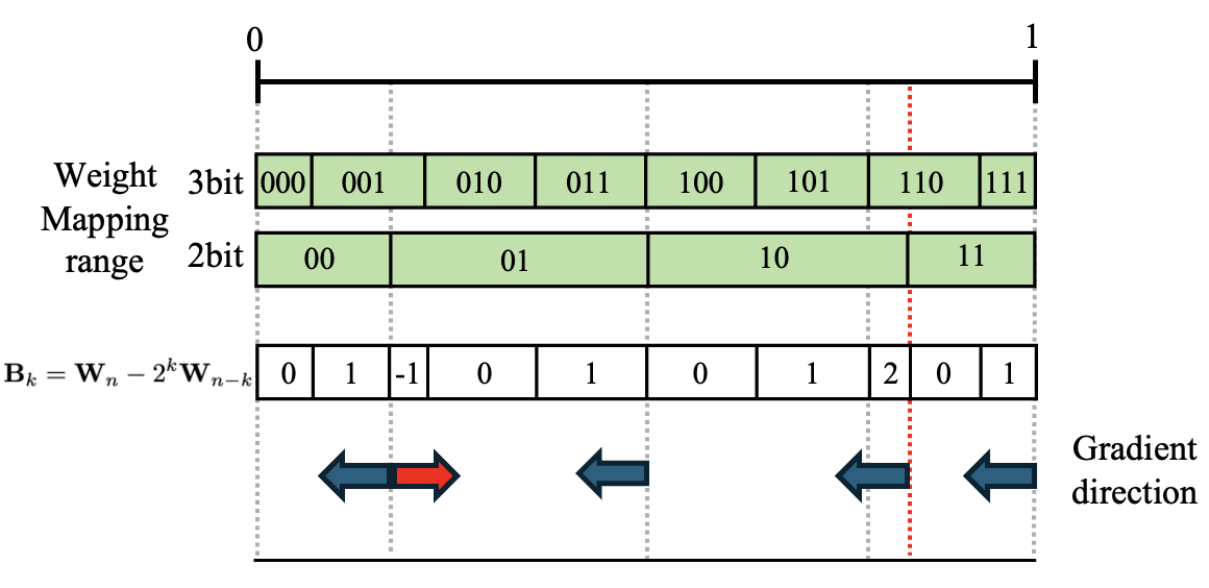}
        \caption{Except for 001, gradient directions of other LSB-nonzero values do not work properly, as they only have one decreasing direction. It even has a gradient for 110, which should not exist.}
        \label{fig:range_a}
    \end{subfigure}
    \vspace{5pt} %
    \begin{subfigure}{1\linewidth}
        \includegraphics[width=\linewidth]{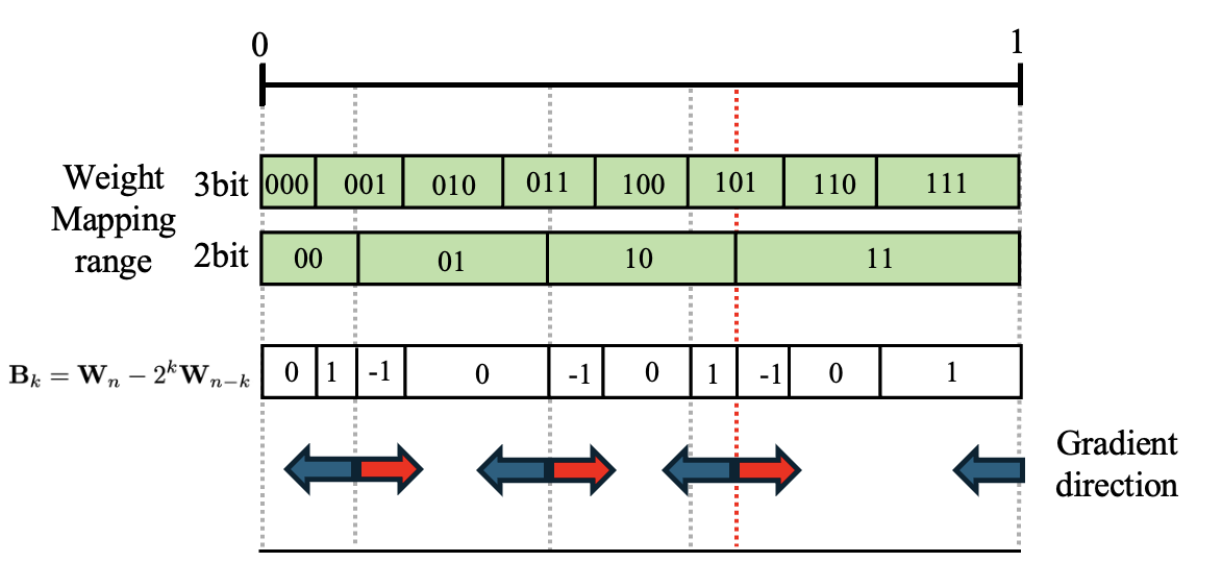}
        \caption{All gradient directions of LSB-nonzero values work well.}
        \label{fig:range_b}
    \end{subfigure}
    \caption{Three-bit and two-bit quantized weight mapping range of (a) DoReFa quantizer and (b) round-clamp quantizer.}
    \label{fig:range}
\end{figure}

To this end, we first propose a bipartite bit-slicing method that can directly compute the values of the LSBs from a model weight element. Specifically, consider a weight element $\mathbf{W}$ scaled to $[0,1]$. We want to quantize $\mathbf{W}$ to a $n$-bit fixed point number $\mathbf{W}_n$, and compute the value represented by the $k$ LSBs of $\mathbf{W}_n$, denoted as $\mathbf{B}_k$. Note that the quantized weight is equivalent to the summation of $\mathbf{B}_k$ and the top $(n-k)$ MSBs of $\mathbf{W}_n$ shifted left by $k$ bits. We therefore propose to apply a quantizer where the top $(n-k)$ MSBs of $\mathbf{W}_n$ is exactly the $(n-k)$-bit quantized $\mathbf{W}_{n-k}$, so that $\mathbf{B}_k$ can be easily computed from $\mathbf{W}$ as
\begin{equation}
\mathbf{B}_k = \mathbf{W}_n - 2^k\mathbf{W}_{n-k}.
\end{equation}
The DoReFa~\cite{zhou2016dorefa} quantizer, as discussed in \cref{eq:ste}, is the most common linear quantizer used in previous work.
However, when observing the $\mathbf{B}_k$ computation results under the DoReFa quantizer, as illustrated in \cref{fig:range}(a), we notice two issues hindering its usage in inducing bit-level. Firstly, the direction of inducing $\mathbf{B}_k$ to be zero is pointing towards the negative direction for most of the values of $\mathbf{W}$. This will induce the value of $\mathbf{W}$ to be constantly smaller as the training proceed, ultimately deviating away from its optimal value. Secondly, due to the use of a scaling factor of $(2^n -1)$ in the rounding function, the quantization bin boundaries under different quantization precision are not well aligned. For example, some values quantized to ``110'' under 3-bit are mapped to ``10'' instead of ``11'' under 2-bit, leading to an error in the LSB computation.

To address this limitation, we introduce the RoundClamp quantizer, a novel quantization scheme designed to effectively compute LSBs for bit-level sparsity exploration. We adjust the scaling factor in the rounding function to be $2^n$ instead of $(2^n -1)$, which leads to the formulation of the quantizer as
\begin{equation}
\mathbf{W}_n = q_r(\mathbf{W}; n) = \frac{1}{2^n-1}\min(\left\lceil 2^n\mathbf{W} \right\rfloor, 2^n-1),
\end{equation}
where $\left\lceil \cdot \right\rfloor$ denotes the rounding function. Note that a clamping is needed to ensure the quantized value stays in the valid range of $[0,1]$.

As illustrated in \cref{fig:range}(b), RoundClamp adjusts the quantization bin boundaries of $(n-1)$-bit quantized weights to align with the midpoint of the quantization bins in $n$-bit quantization. This ensures that for those weights where the quantized value having nonzero LSBs, they have the chances to be round both up or down to the nearest bin with zero LSBs. When sparsity-inducing regularizer is applied to the computed LSBs to induce sparsity through training, we observe that the gradient directed by the RoundClamp quantizer can lead to zero LSBs and precision reduction more effectively. 
  
\begin{figure}[tb]
    \centering
    \begin{subfigure}{0.45\linewidth}
        \includegraphics[width=\linewidth]{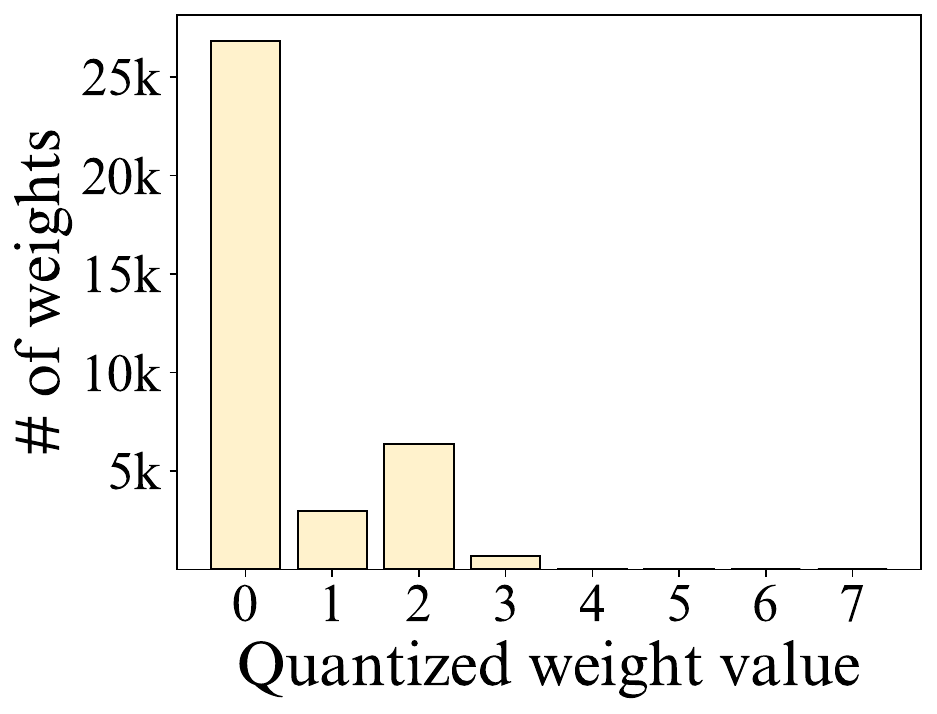}
        \caption{DoReFa quantizer}
        \label{fig:weight_dist_a}
    \end{subfigure}
    \hfill
    \begin{subfigure}{0.45\linewidth}
        \includegraphics[width=\linewidth]{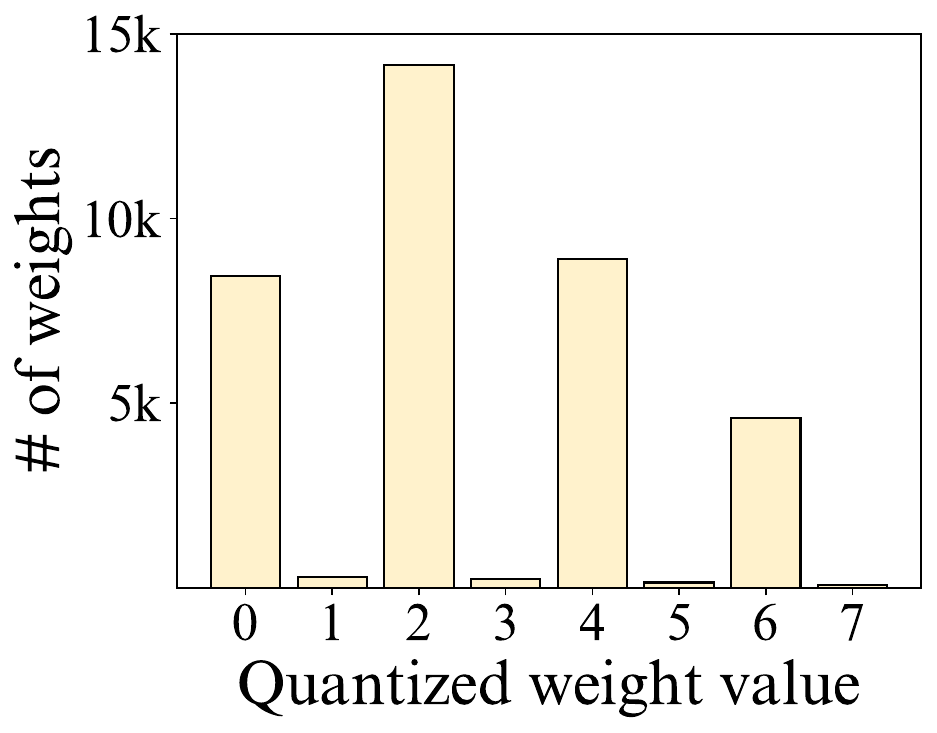}
        \caption{Round-clamp quantizer}
        \label{fig:weight_dist_b}
    \end{subfigure}
    
    \caption{Weight distribution of training with (a) DoReFa and (b) round-clamp quantizer. Both weight distributions are the results of 179 epochs, right before pruning.}
    \label{fig:weight_dist}
\end{figure}

The difference in the behavior of the two quantizers is evident in the resulting weight distributions after training, as shown in \cref{fig:weight_dist}. In the case of DoReFa~\cite{zhou2016dorefa}, depicted in \cref{fig:weight_dist}(a), the weight distribution exhibits pronounced spikes at zero. This outcome arises due to the dominance of the negative gradient in the process of sparsifying the nonzero LSBs. In contrast, the RoundClamp quantizer produces a weight distribution characterized by higher densities at LSB-zero positions and lower densities at LSB-nonzero positions. This result aligns with the intended design, effectively driving the weights toward near-zero LSB values, as desired.
%%%%%%%%%%%%%%%%%%%%%%%%%%%%%%%%%%%%%%%%%%%%%%%%%%%%%%%%%%%%%%%%%%%%%%%%%%%%%%

\noindent\textbf{LSBs Computation and Regularization.} With the RoundClamp quantizer, we can compute the LSB of floating-point weight $\mathbf{W}$ under $n$-bit quantization as
\begin{equation}
\mathbf{B}_k = \mathbf{W} - 2^k q_r(\mathbf{W}; n-k).
\end{equation}
$\mathbf{B}_k$ is a continuous function with respect to the weight $\mathbf{W}$. Through this process, the LSB can be extracted without bit-level splitting.
%%%%%%%%%%%%%%%%%%%%%%%%%%%%%%%%%%%%%%%%%%%%%%%%%%%%%%%%%%%%%%%%%%%%%%%%%%%%%%
%%%%%%%%%%%%%%%%%%%%%%%%%%%%%%%%%%%%%%%%%%%%%%%%%%%%%%%%%%%%%%%%%%%%%%%%%%%%%%

To enforce LSB sparsity and precision reduction, we apply the $\ell_1$ regularization to the computed LSB across all layers during training. The regularization term is expressed as:
\begin{equation}
\text{R}(\mathbf{B}_k) = \sum_{\forall l} \lvert\mathbf{B}^{(l)}_k\rvert, \end{equation}
where $\mathbf{B}^{(l)}_k$ denotes the LSBs of the weights for layer $l$, and $\lvert\cdot\rvert$ represents the absolute value function. 
The gradient of the regularization loss with respect to $\mathbf{W}$ is given by:
\begin{equation}
\frac{\partial{\text{R}(\mathbf{B_k})}}{\partial{\mathbf{W}}} = \text{sign}(\mathbf{B_k}),
\end{equation}
which will guide the weight towards the nearest low-precision value following the RoundClamp quantizer design.

The overall quantization-aware training objective with the $\ell_1$ regularization is defined as:
\begin{equation} \label{eq:total_loss}
\min_{\mathbf{W}} \ \mathcal{L}(\mathbf{W}_n) + \lambda\sum_{\forall l} \lvert\mathbf{B}_k^{(l)}\rvert,
\end{equation}
where $\mathcal{L}(\cdot)$ is the original training loss, $\mathbf{W}_n$ is the $n$-bit quantized weight with the RoundClamp quantizer, and $\lambda$ is the regularization strength trading off model compression rate and model performance. The objective can be optimized via gradient-based optimizer with the help of the straight-through estimator.

\subsection{Hessian-aware Aggressive Pruning }
\label{sec:haap}
%%%%%%%%%%%%%%%%%%%%%%%%%%%%%%%%%%%%%%%%%%%%%%%%%%%%%%%%%%%%
\begin{figure}[tb]
    \centering
    \begin{subfigure}{0.48\textwidth}
        \centering
        \includegraphics[width=\linewidth]{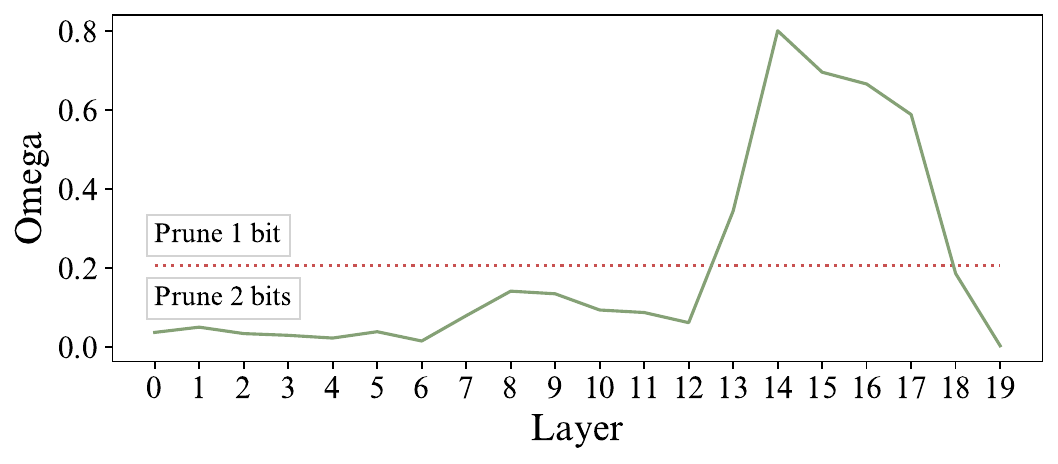}
        \caption{Omega values during the first pruning step.}
        \label{fig:hessian_first}
    \end{subfigure}
    \hfill
    \begin{subfigure}{0.48\textwidth}
        \centering
        \includegraphics[width=\linewidth]{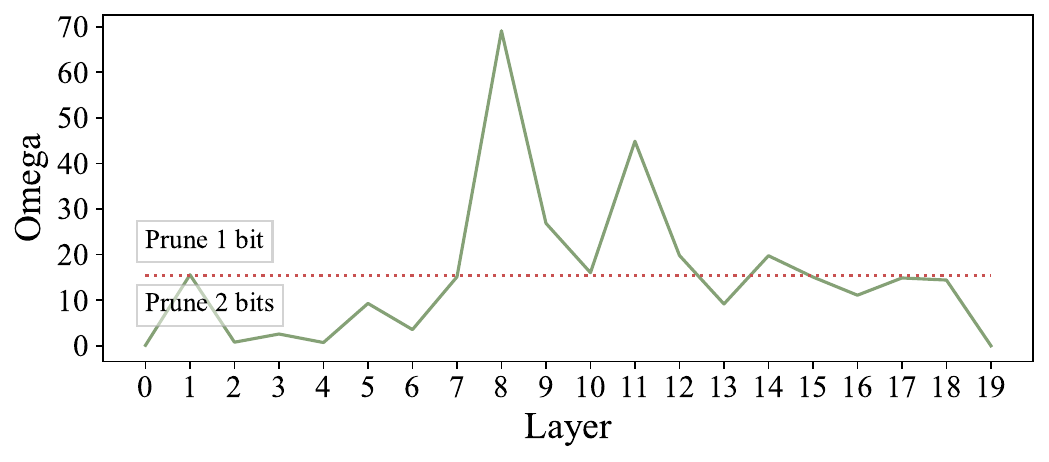}
        \caption{Omega values during the last pruning step.}
        \label{fig:hessian_last}
    \end{subfigure}
    \caption{Comparison of Omega values for each layer of ResNet-20 between the first and last pruning steps. Layers with omega values exceeding the average (red dotted line) are pruned by 1 bit, while those below are pruned by 2 bits.}
    \label{fig:hessian_omega}
\end{figure}
%%%%%%%%%%%%%%%%%%%%%%%%%%%%%%%%%%%%%%%%%%%%%%%%%%%%%%%%%%%%
\begin{algorithm}[!t] 
    \caption{Overall training algorithm}
    \label{al:overall}
    \textbf{Input:} Data \textbf{X}, label \textbf{Y}\\
    \textbf{Output:} mixed-precision model G
    \begin{algorithmic}[1]
    \State Initialize: $p$, $q$ in G
    \State Initialize: regularization strength $\lambda$, pruning interval $I$, pruning threshold $\alpha$, target compression $\Gamma$
    \State Initialize: LSB-nonzero rate $\beta$, compression $\gamma$
    \For{epoch = 1, ..., T}
        \For{batch from \textbf{X}, \textbf{Y}}
            \State Compute forward pass of G
            \State \textit{\color{teal}\# LSB $L_1$ regularization}
            \State Update parameters with $ L = L_{ce} + \lambda\text{R}(\mathbf{B})$ 
        \EndFor
        \If{epoch \% $I$ == 0 \textbf{and} $\gamma > \Gamma$}
            \State Calculate Hessian trace $Tr(H)$
            \State Initialize $\Omega$
            \State \textit{\color{teal}\# Omega and LSB-nonzero rate calculation}
            \For{Quantized layer $l$ in G}
                \State $\Omega_l \leftarrow Tr(H_l)||\mathbf{W}^{(l)}_{q_l} - \mathbf{W}^{(l)}||^2$
                \State $\beta_l \leftarrow \text{sum}(\mathbf{B} > 2^{p_l-1})$
            \EndFor
            \State \textit{\color{teal}\# Pruning}
            \State Ascending sort $\beta$
            \For{Quantized layer $l$ in G}
                \If {$\beta_l < \alpha$} 
                    \State $q_l = q_l - p_l$
                \EndIf
    
                \If {$\gamma > \Gamma$} 
                    \State break
                \EndIf
            \EndFor
            
            \State \textit{\color{teal}\# Hessian aware changing pruning bit}
            \For{Quantized layer $l$ in G}
                \If {$\Omega_l < mean(\Omega)$} 
                    \State $p_l = 2$
                \Else
                    \State $p_l = 1$
                \EndIf
            \EndFor      
        \EndIf
    \EndFor
    \end{algorithmic}
\end{algorithm}
The bipartite bit slicing introduces a new hyperparameter, $k$, to the MSQ process. As the MSQ training objective in \cref{eq:total_loss} aims to sparsify the $k$ LSBs in each layer, $k$ decides each layer's precision reduction speed. Intuitively, if a layer's weight can tolerate more perturbations without impacting the final model loss, i.e. less sensitive, it's precision can be reduced at a higher pace. On the other hand, precision reduction on a sensitive layer should be cautious to avoid catastrophically hurting the model performance.  

To incorporate sensitivity information into the pruning process, we propose Hessian-aware Aggressive Pruning, where we measure the sensitivity of each layer using their Hessian statistics and assign a larger $k$ to those layers that are less sensitive. 
Specifically, we follow the methodology proposed in HAWQ-V2~\cite{dong2020hawq}, which identifies the Hessian trace can be used as a reliable sensitivity metric for quantized models. The sensitivity of each layer, $\Omega_l$, is calculated as:
\begin{equation}
\Omega_l = Tr(H_l)\|\mathbf{W}^{(l)}_n - \mathbf{W}^{(l)}\|^2,
\end{equation}
where $l$ refers to the $l^{th}$ layer, $Tr(H_i)$ denotes Hessian trace with respect to the $l^{th}$ layer weight, $\mathbf{W}_n$ represents the quantized weight under the current precision, and $\|\cdot\|$ the $L_2$ norm. 

MSQ uses a heuristic-based thresholding method to decide the precision reduction speed $k$ for each layer. We set the threshold to be the averaged sensitivity of all layers.
If the sensitivity of a layer is larger than the threshold, we set $k=1$, where as layers with sensitivity lower than the threshold use $k=2$, as shown in \cref{fig:hessian_omega}. Additionally, as the training and LSB pruning step progresses, the sensitivity metric $\Omega_l$ for each layer will be recomputed to capture its change, which can be observed in the transition from \cref{fig:hessian_first} to \cref{fig:hessian_last}.
This adaptive strategy allows for accelerated training by pruning multiple bits at once in low-sensitivity layers while preserving accuracy in high-sensitivity layers. Moreover, it ensures that Hessian information is effectively utilized to optimize the pruning process, resulting in both computational efficiency and improved performance in quantized networks.
\begin{figure*}[tb]
  \centering
  \begin{subfigure}[t]{0.33\linewidth}
    \centering
    \includegraphics[width=\linewidth]{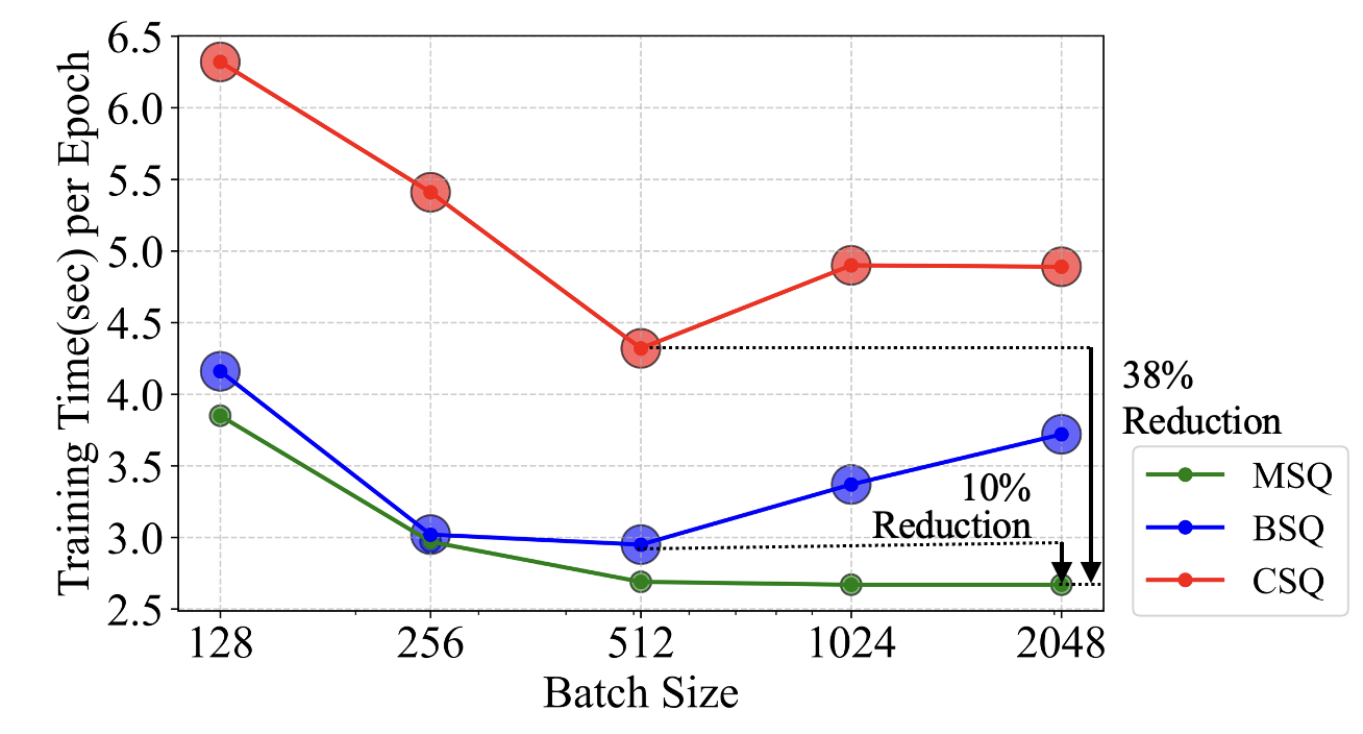}
    \caption{ResNet-20}
    \label{fig:resnet20}
  \end{subfigure}
  \hfill
  \begin{subfigure}[t]{0.33\linewidth}
    \centering
    \includegraphics[width=\linewidth]{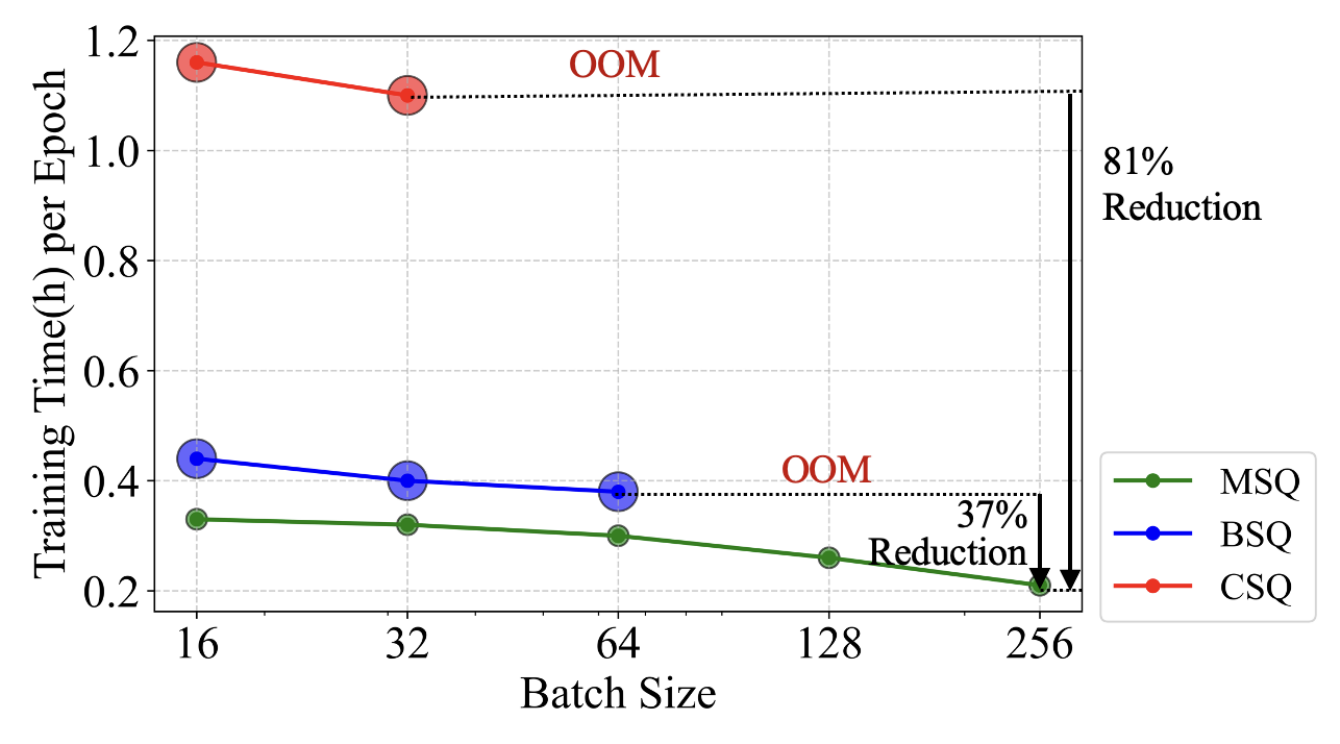}
    \caption{ResNet-18}
    \label{fig:resnet18}
  \end{subfigure}
  \hfill
  \begin{subfigure}[t]{0.33\linewidth}
    \centering
    \includegraphics[width=\linewidth]{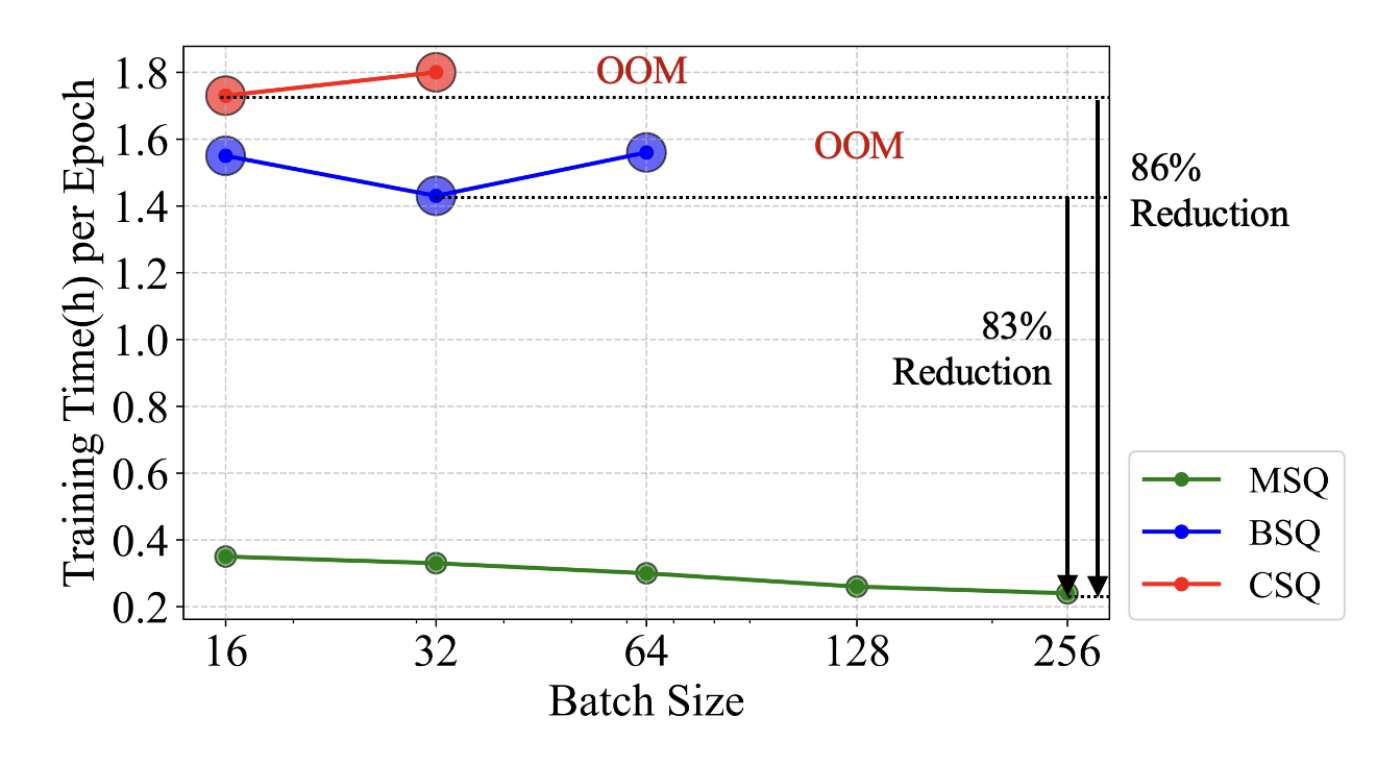}
    \caption{ResNet-50}
    \label{fig:resnet50}
  \end{subfigure}
  \caption{Comparison of training time per epoch for different methods across varying batch sizes. Results are reported up to the maximum batch size that does not cause out-of-memory (OOM). Circle sizes represent the number of trainable parameters.}
  \label{fig:training_time_batch_size}
\end{figure*}

\subsection{Overall Training Algorithm}
\label{sec:ota}
The complete training workflow is summarized in \Cref{al:overall}. The model is trained using the objective defined in \cref{eq:total_loss}, enabling simultaneous weight optimization and LSB sparsification through $L_1$ regularization. At every pruning interval, the sensitivity metric $\Omega$ is calculated for all layers, guiding the adjustment of the pruning bit amount $p$ for the subsequent pruning step.

The pruning strength is controlled by two key parameters: the regularization strength $\lambda$ and the pruning threshold $\alpha$,which are applied uniformly across all layers. Pruning occurs when the LSB nonzero rate $\beta$ for a given layer falls below the pruning threshold $\alpha$. 
The pruning continues until the target compression ratio $\Gamma$ is achieved. In the final round of pruning, we sort $\beta$ to prioritize pruning layers with the lowest LSB nonzero rates until the target model compression ratio is reached. Once the $\Gamma$ is reached, regularization and pruning are completely stopped, and training continues as a standard QAT process to improve model performance under the finalized quantization scheme.

\section {Evaluation}
\label{sec:evaluation}

We evaluate MSQ on a variety of neural network architectures to demonstrate its generality and effectiveness. For convolutional networks, we test MSQ on ResNet-20~\cite{he2016deep} using the CIFAR-10 dataset~\cite{krizhevsky2009learning}, and on ResNet-18 and ResNet-50~\cite{he2016deep} using the ImageNet dataset~\cite{deng2009imagenet}. To further assess generalization to heterogeneous CNN architectures, we also evaluate MSQ on MobileNetV3-Large ~\cite{howard2019mobilenetv3}, which include depthwise convolutions and squeeze-and-excitation blocks.
For transformer-based architectures, we assess MSQ on ImageNet using DeiT~\cite{touvron2021deit} and Swin-T~\cite{liu2021swin}, representing both lightweight and hierarchical Vision Transformers (ViTs).
For CNNs, we compare MSQ with existing uniform quantization methods~\cite{zhou2016dorefa,choi2018pact} as well as mixed-precision quantization approaches~\cite{dong2020hawq,yao2021hawqv3,yang2021bsq,xiao2023csq}. For ViTs, we evaluate MSQ against various ViT-specific quantization techniques~\cite{esser2019lssq,li2022qvit,navin2025vit,touvron2021deit}.
\begin{table}[b]
        \centering
        \caption{Training resource usage for each quantization method.}
        \vspace{-1em}
        \label{table:training}
        \small
        \scriptsize  
        \begin{tabular}{l|c|>{\centering\arraybackslash}p{1.05cm}>{\centering\arraybackslash}p{1.05cm}>{\centering\arraybackslash\columncolor{gray!20}}p{1.05cm}}
            \toprule
            % \multicolumn{2}{c|}{} & \multicolumn{3}{c}{Method} \\
            % \cmidrule{3-5}
            \multicolumn{2}{l|}{Network} & BSQ & CSQ & \textbf{MSQ} \\
            \midrule
            \multirow{5}{*}{ResNet-20}
                & Epochs                    & 350      & 600      & 400 \\
                & Train Batch Size          & 256      & 256      & 1024 \\
                & Total Time (h)          & 0.36     & 0.90     & \textbf{0.33} \\
                & Params (M)   & 2.16  & 2.16  & \textbf{0.27} \\

                & Peak Memory(GB)               & 2.21       & 2.40      & \textbf{2.77} \\
            \midrule
            \multirow{5}{*}{ResNet-18}
                & Epochs                    & 90       & 200      & 100 \\
                & Train Batch Size          & 64       & 32       & 256 \\
                & Total Time (h)            & 63.92    & 220.32   & \textbf{21.11} \\
                & Params (M)   & 93.52  & 93.52  & \textbf{11.69} \\

                & Peak Memory(GB)               &  10.13      & 10.41       & \textbf{11.13} \\
            \midrule
            \multirow{5}{*}{ResNet-50}
                & Epochs                    & 90       & 200      & 100 \\
                & Train Batch Size          & 32       & 16       & 256 \\
                & Total Time (h)            & 128.76   & 346.12   & \textbf{24.30} \\
                & Params (M)   & 204.8  & 204.8  & \textbf{25.6} \\

                & Peak Memory(GB)               & 11.54       & 12.82       & \textbf{12.22} \\
            \bottomrule
        \end{tabular}
    \end{table}
\subsection{Experimental Setup}
We employ a consistent set of hyperparameters for experiments conducted on the same model.
Models trained on CIFAR-10 use the SGD optimizer with an initial learning rate of 0.1 and a warm-start cosine annealing scheduler. For ImageNet, we use the SGD optimizer with an initial learning rate of 0.01, also adopting a warm-start cosine annealing schedule.
Experiments on CIFAR-10 are trained from scratch and for 400 epochs, whereas on ImageNet, we use floating-point pretrained models and train for 100 epochs.  
These training durations are comparable to the total epochs used in prior methods, such as BSQ ~\cite{yang2021bsq} and CSQ ~\cite{xiao2023csq}. 
For activation quantization, we use uniform quantization and report the bit precision in the “A-Bits" column. In ImageNet experiments, all activations remain at full precision, while in ViT experiments, activations are quantized to 8-bit. 

\subsection{Experimental Results}

\begin{table}[t]
    \centering
    \setlength{\tabcolsep}{2pt}
    \scriptsize
    \caption{Quantization results of ResNet-20 models on the CIFAR-10 dataset.}
    \vspace{-1em}
    \label{table:cifar10}
    \begin{tabular}{
        >{\centering\arraybackslash}p{1.1cm}  % Method
        |>{\centering\arraybackslash}p{0.64cm} >{\centering\arraybackslash}p{0.64cm} >{\centering\arraybackslash}p{0.64cm}  % ABit 32
        |>{\centering\arraybackslash}p{0.64cm} >{\centering\arraybackslash}p{0.64cm} >{\centering\arraybackslash}p{0.64cm}  % ABit 4
        |>{\centering\arraybackslash}p{0.64cm} >{\centering\arraybackslash}p{0.64cm} >{\centering\arraybackslash}p{0.64cm}  % ABit 3
        }
        \toprule
        \multicolumn{1}{c|}{A-Bits} 
        & \multicolumn{3}{c|}{32} 
        & \multicolumn{3}{c|}{3} 
        & \multicolumn{3}{c}{2} \\
        \midrule
        Method 
        & W-Bit & Comp & Acc
        & W-Bit & Comp & Acc
        & W-Bit & Comp & Acc \\
        \midrule
        FP        & 32 & 1.00  & 92.62 & --  & --    & --    & --  & --    & -- \\
        LQ-Nets   & 3  & 10.67 & 92.00 & 3   & 10.67 & 91.60 & 2  & 16.00 & 90.20 \\
        PACT      & -- & --    & --    & 3   & 10.67 & 91.60 & 2  & 16.00 & 89.70 \\
        DoReFa    & -- & --    & --    & 3   & 10.67 & 89.90 & 2  & 16.00 & 88.20 \\
        BSQ       & MP & 19.24 & 91.87 & MP  & 11.04 & 92.16 & MP & 18.85 & 90.19 \\
        CSQ T2    & MP & 16.00 & 92.68 & MP  & 16.93 & 92.14 & MP & 16.41 & 90.33 \\
        \rowcolor{gray!20}
        \textbf{MSQ} & MP & \textbf{16.13} & \textbf{92.17} 
                    & MP & \textbf{17.43} & \textbf{92.00} 
                    & MP & \textbf{19.13} & \textbf{90.22} \\
        \rowcolor{gray!20}
        \textbf{MSQ} & MP & \textbf{20.13} & \textbf{92.15} 
                    & -- & -- & -- 
                    & MP & \textbf{16.43} & \textbf{90.60} \\
        \bottomrule
    \end{tabular}
    \vspace{-0.5em}
\end{table}
\begin{table}[tb]
    \centering
    \setlength{\tabcolsep}{0pt}
    \small
    \caption{Quantization results of ResNet-18 and ResNet-50 models on the ImageNet dataset.}
    \label{table:imagenet}
    \begin{tabular}{l|ccc|ccc}
        \toprule
        
        \multicolumn{2}{c}{}            & ResNet-18         & \multicolumn{2}{c}{}              & ResNet-50         &                   \\
        \midrule
        Method          & W-Bits        & Comp(x)           & Acc(\%)           & W-Bits        & Comp(x)           & Acc(\%)           \\
        \midrule
        FP              & 32            & 1.00              & 69.76             & 32            & 1.00              & 76.13             \\
        \midrule
        DoReFa~\cite{zhou2016dorefa}           & 4             & 6.40              & 68.4              & 3             & 10.67             & 69.90             \\
        PACT~\cite{choi2018pact}             & 4             & 8.00              & 69.2              & 3             & 10.67             & 75.30             \\
        LQ-Nets~\cite{zhang2018lq}          & 3             & 10.67             & 69.30             & 3             & 10.67             & 74.20             \\
        HAWQ-V3~\cite{yao2021hawqv3}         & 4             & 8.00              & 68.45             & 4             & 8.00              & 74.24             \\
        HAQ~\cite{wang2019haq}             & -             & -                 & -                 & MP            & 10.57             & 75.30             \\
        BSQ~\cite{yang2021bsq}             & -             & -                 & -                 & MP            & 13.90             & 75.16             \\
        CSQ T3~\cite{xiao2023csq}          & MP            & 10.67             & 69.73             & MP            & 10.67             & 75.47             \\
        \rowcolor{gray!20}
        \textbf{MSQ}    & \textbf{MP}   & \textbf{11.84}        & \textbf{69.74}        & \textbf{MP}   & \textbf{10.89}        & \textbf{75.32}    \\
        \bottomrule
    \end{tabular}
\end{table}

We compare MSQ with previous baseline quantization methods. 
In all tables, “FP” refers to the full-precision model, and “MP” denotes mixed-precision weight quantization. The weight compression ratio, “Comp" is calculated relative to the full-precision model. 
The target compression ratios of 16.00 and 10.67 correspond approximately to average weight bit-widths of 2 and 3 bits, respectively.

\noindent\textbf{Training Efficiency Comparison.}
As shown in Fig.\ref{fig:training_time_batch_size}, This comparison presents the training time for each batch size across different methods (BSQ, CSQ, and MSQ). The batch size for each method was increased until it reached the out-of-memory limit. The specific details are summarized in Table \ref{table:training}, where the training resource usage is reported. MSQ achieves up to 8× fewer trainable parameters compared to BSQ and CSQ, primarily due to its structural design that avoids explicit bit-level splitting during training. Consequently, MSQ allows for larger batch sizes before reaching the out-of-memory limit, ultimately reducing training time.
To ensure a fair comparison, we also measured the total training time using batch sizes adjusted to yield similar peak memory usage for each method on an RTX GeForce 4080 Super. As the number of parameters increased, MSQ’s efficiency advantage became more significant. In particular, on ResNet-50, MSQ achieved a 5.3× speedup over BSQ and a 14.2× speedup over CSQ. This acceleration stems from MSQ’s architectural efficiency, which enhances parallelism and minimizes computational overhead by eliminating bit-level splitting during training.

\noindent\textbf{ResNet Results.}
We evaluate MSQ on ResNet-20 with CIFAR-10, while ImageNet is used for ResNet-18 and ResNet-50. As shown in \Cref{table:cifar10}, MSQ achieves higher compression ratios than CSQ under 32-bit and 3-bit activation settings, with minimal accuracy drop. In the 2-bit case, MSQ even improves accuracy over CSQ. On ImageNet (\Cref{table:imagenet}), MSQ outperforms prior methods on ResNet-18 and achieves comparable performance to CSQ on ResNet-50.

\noindent\textbf{MobileNetV3 Results.}
To assess the generalization capability of MSQ, we evaluate it on MobileNetV3-Large using the ImageNet dataset. This architecture includes diverse components such as depthwise separable convolutions and squeeze-and-excitation (SE) blocks, which present additional challenges for quantization. As shown in \Cref{table:mobilenet}, MSQ achieves higher compression ratios compared to prior methods while preserving model accuracy. 

\noindent\textbf{ViT Results.}  
Table~\ref{table:vit} presents the MSQ results for Vision Transformer models on ImageNet datasets, including DeiT \cite{touvron2021deit} and Swin \cite{liu2021swin}. In these experiments, we use the 4-bit quantized checkpoint from the OFQ \cite{liu2023ofq} repository as the pretrained model and fine-tune it using using MSQ. MSQ achieves a better efficiency-accuracy trade-off compared to previous methods, delivering comparable performance at higher compression ratios. These results demonstrate that MSQ is effective for transformer-based models, extending its applicability beyond CNN architectures. 

\begin{table}[t]
        \centering
        \setlength{\tabcolsep}{0pt}
        \scriptsize
        \caption{Comparison of quantization methods on DeiT-T/S and Swin-T with compression ratio and Top-1 accuracy on ImageNet. }
        \vspace{-1em}
        \label{table:vit}
        \begin{tabular}{
            >{\centering\arraybackslash}p{1.2cm}
            |>{\centering\arraybackslash}p{0.9cm}
            |>{\centering\arraybackslash}p{1.0cm}
              >{\centering\arraybackslash}p{1.0cm}
            |>{\centering\arraybackslash}p{1.0cm}
              >{\centering\arraybackslash}p{1.0cm}
            |>{\centering\arraybackslash}p{1.0cm}
              >{\centering\arraybackslash}p{1.0cm}}
            \toprule
            \multicolumn{2}{c|}{} & \multicolumn{2}{c|}{DeiT-T} & \multicolumn{2}{c|}{DeiT-S} & \multicolumn{2}{c}{Swin-T} \\
            \midrule
            Method & W-Bits & Comp(×) & Acc(\%) & Comp(×) & Acc(\%) & Comp(×) & Acc(\%) \\
            \midrule
            LSQ          & 3  & 10.67 & 68.09 & 10.67 & 77.76 & 10.67 & 78.96 \\
            Mix-QViT     & 3  & 10.67 & 69.62 & 10.67 & 78.08 & 10.67 & 79.45 \\
            QViT         & 3  & 10.67 & 67.12 & 10.67 & 78.45 & 10.67 & 80.06 \\
            OFQ          & 3  & 10.67 & 72.72 & 10.67 & 79.57 & 10.67 & 79.57 \\
            OFQ          & 4  & 8.00  & 75.46 & 8.00  & 81.10 & 8.00  & 81.88 \\
            \rowcolor{gray!20}
            \textbf{MSQ} & MP & \textbf{10.54} & \textbf{74.74} & \textbf{9.58} & \textbf{80.64} & \textbf{9.14} & \textbf{81.38} \\
            \bottomrule
        \end{tabular}
    \end{table}

\begin{table}[b]
    \centering
    \setlength{\tabcolsep}{4pt}
    \small
    \caption{Quantization results of MobileNetV3-L on the ImageNet dataset.}
    \label{table:mobilenet}
    \begin{tabular}{l|ccc}
        \toprule
        Method & W-Bits & Comp($\times$) & Acc(\%) \\
        \midrule
        FP & 32 & 1.00 & 75.27 \\
        \midrule
        DoReFa~\cite{zhou2016dorefa} & 8 & 4.00 & 74.44 \\
        \rowcolor{gray!20}
        \textbf{MSQ} & \textbf{MP} & \textbf{5.36} & \textbf{74.29} \\
        \midrule
        DoReFa~\cite{zhou2016dorefa} & 4 & 8.00 & 72.92 \\
        \rowcolor{gray!20}
        \textbf{MSQ} & \textbf{MP} & \textbf{10.30} & \textbf{73.58} \\
        \bottomrule
    \end{tabular}
\end{table}

\subsection{Ablation Study}
In this section, we delve into the critical design aspects of the MSQ, particularly examining the influence of the Hessian information in determining the bit scheme for mixed-precision quantization. The experiments used in this section are performed using ResNet-20 models~\cite{he2016deep}, leveraging 3-bit activation quantization, and are evaluated on the CIFAR-10 dataset~\cite{krizhevsky2009learning}.

\begin{figure}[t]
\centerline{\includegraphics[width=\columnwidth]{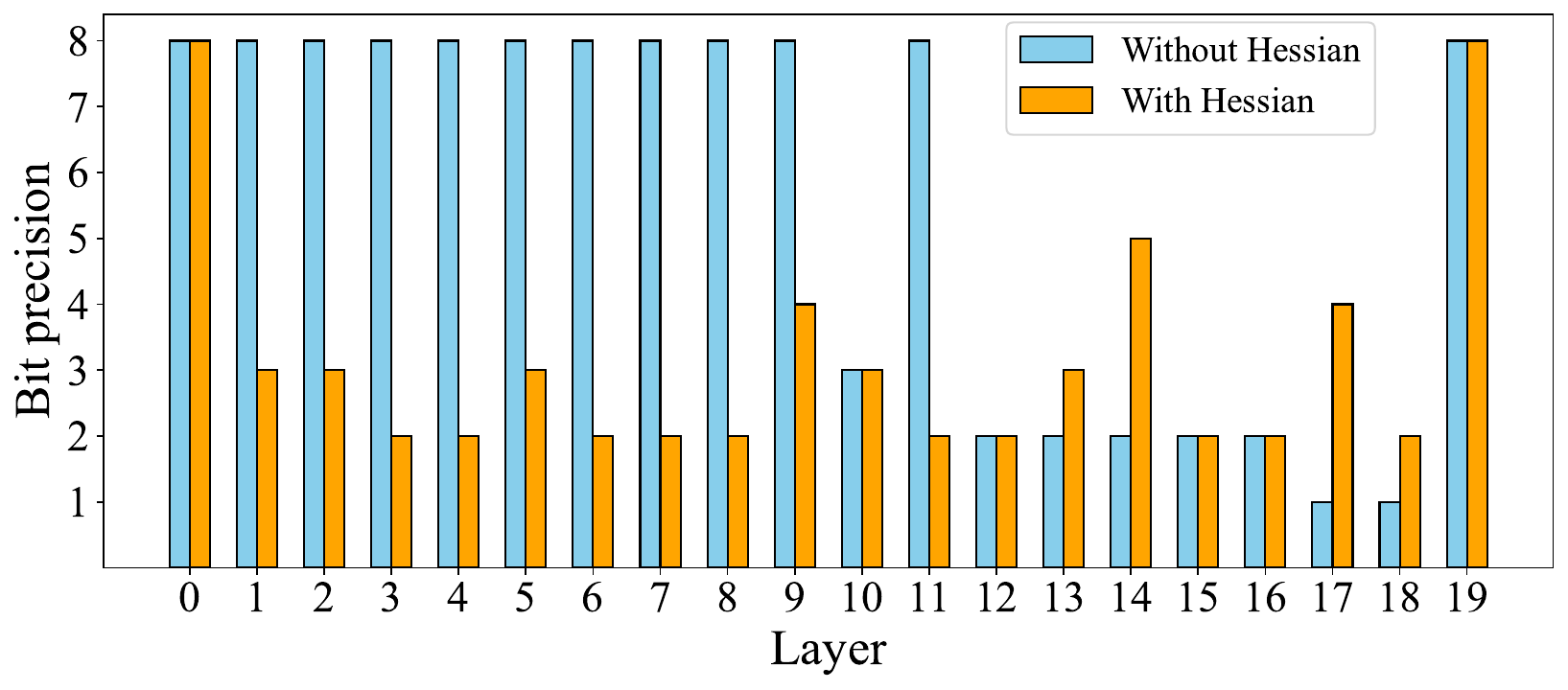}}
\caption{Final bit scheme selected during the training process. The bit scheme is determined at epoch 210 for the case without Hessian (accuracy: 91.23\%), and at epoch 150 for the case with Hessian (accuracy: 91.93\%).}
\label{fig:E_bit_scheme}
\vspace{-10pt}
\end{figure}

\begin{figure}[tb]
\centerline{\includegraphics[width=\columnwidth]{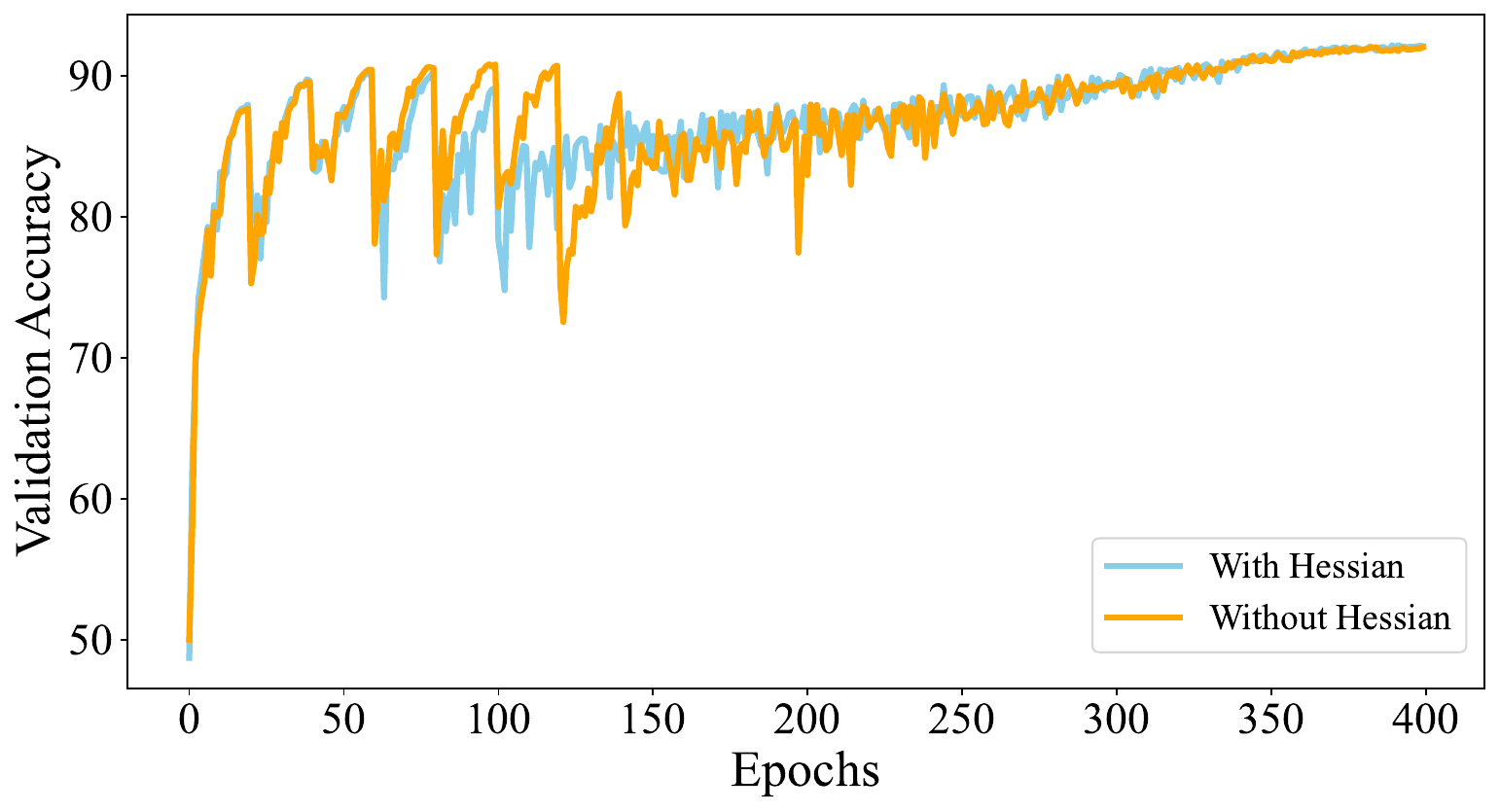}}
\caption{The change of validation accuracy with and without Hessian during training. Accuracy drop is observed when LSB pruning happens.}
\label{fig:val_prec}
\vspace{-10pt}
\end{figure}
\newpage
\noindent\textbf{Effectiveness of Hessian.} 
As discussed in Sec.~\ref{sec:haap}, MSQ leverages Hessian information to account for layer sensitivity, enhancing the efficiency of the bit-pruning process. Fig.~\ref{fig:E_bit_scheme} compares the final bit schemes after training with and without Hessian guidance. 
Without the Hessian guidance, some layers are constantly pruned while others are less changed, leading to the need of using more pruning epochs to reach the target compression rate and poorer final accuracy. On the other hand, the proposed Hessian-aware aggressive pruning scheme reduces the pruning speed for overly-pruned layers, which becomes more sensitive as they are pruned, while promoting faster pruning of other less sensitive layers. This leads to a more well-behaving quantization scheme. 
As we achieve faster pruning speed with the Hessian guidance, the accuracy drop suffered by the model from each pruning step is smaller as shown in Fig.~\ref{fig:val_prec}, which helps achieving a better compression-accuracy tradeoff with less training.

\noindent\textbf{Quantization Scheme Comparison with BSQ.}
\begin{figure}[tb]
\centerline{\includegraphics[width=\columnwidth]{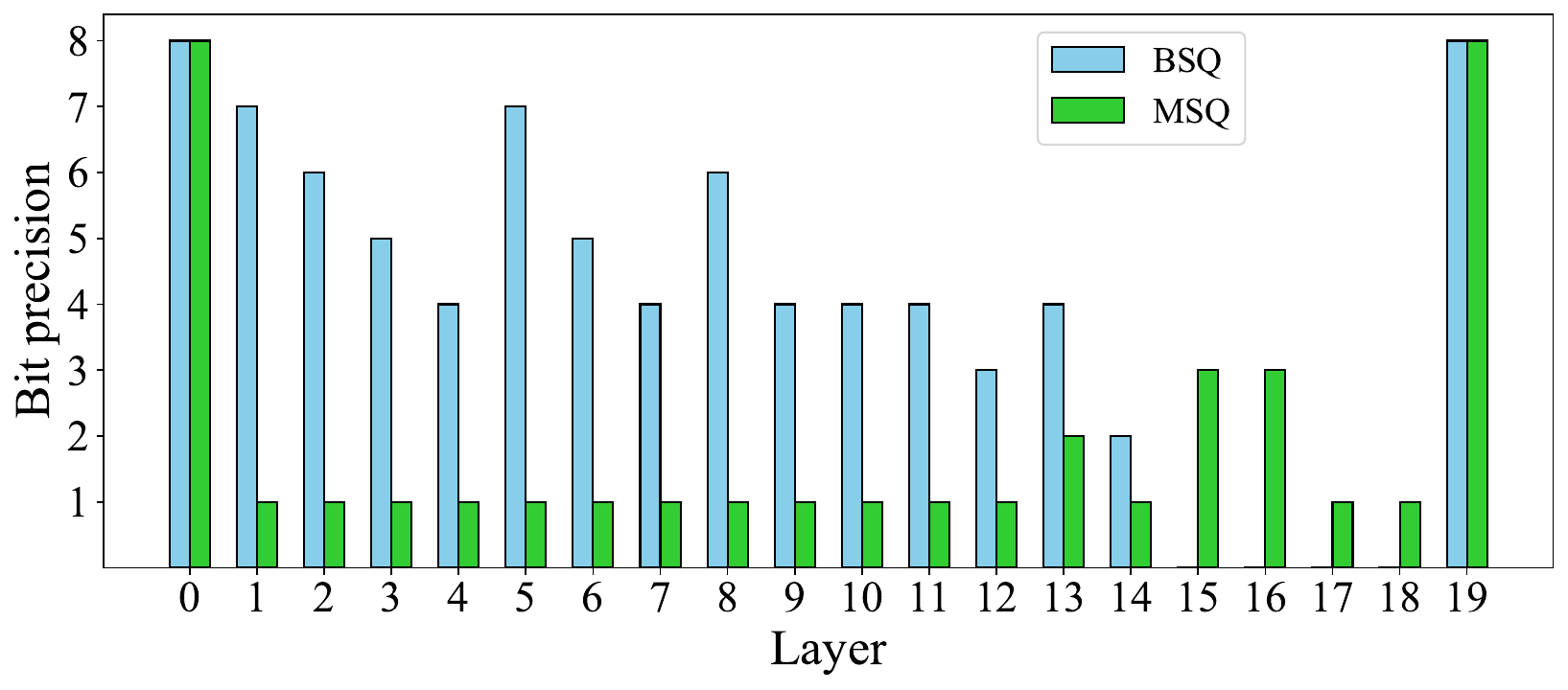}}
\caption{ Comparison of Final Bit Schemes Between MSQ and BSQ During Training: The BSQ's compression ratio is 19.24x with accuracy 91.87\%, while MSQ's compression ratio is 20.13x with accuracy 92.15\%.}
\label{fig:BSQ}
\vspace{-10pt}
\end{figure}
We showcase the quantization scheme achieved by MSQ and BSQ on the ResNet-20 model in Fig.~\ref{fig:BSQ}. We noticed that the sparsity observed by BSQ is more concentrated on a few layers, whereas that achieved by MSQ appears more even. We believe that this difference is induced by the overly aggressive bit-level training and regularization introduced by BSQ, where each bit is trained and regularized separately. In BSQ, strong regularization may reduce some layers to 0-bit precision, effectively removing all weights and allowing the layer to be skipped. MSQ, on the other hand, single out only the LSBs with the guidance of Hessian information, which leads to a better quantization scheme with improved accuracy and compression rate.

\section {Conclusion}
\label{sec:conclusion}
In this work, we introduce MSQ, a novel memory-efficient approach for bit-level mixed-precision quantization DNN training. MSQ computes and regularizes LSBs directly from the trainable parameters, leading to bit-level sparsity and precision reduction without explicit bit splitting. Hessian information is incorporated to guide proper precision reduction rate across the layers. 
Experimental results demonstrate the effectiveness of MSQ, achieving accuracy on par with existing mixed-precision methods while significantly reducing computational resources, including training time and GPU memory usage. This efficiency makes it feasible to train large-scale networks, such as Vision Transformers (ViTs), using a bit-level training approach.

\section*{Acknowledgments}
%%%%%%%%%%%%%%%%%% SKKU %%%%%%%%%%%%%%%%%%%
% 신진 
This work was partly supported by the National Research Foundation of Korea (NRF) grant (No. RS-2024-00345732);
the Institute for Information \& Communications Technology Planning \& Evaluation (IITP) grants (
RS-2020-II201821, 
RS-2019-II190421, 
RS-2021-II212068,
RS-2025-10692981); 
the Technology Innovation Program (RS-2023-00235718, 23040-15FC) funded by the Ministry of Trade, Industry \& Energy (MOTIE, Korea) grant (1415187505); 
the BK21-FOUR Project; 
%%%%%%%%%%%%%%%%%% UofA %%%%%%%%%%%%%%%%%%%
and High Performance Computing (HPC) resources supported by the University of Arizona TRIF, UITS, and Research, Innovation, and Impact (RII), and maintained by the UArizona Research Technologies department.

{
    \small
    \bibliographystyle{ieeenat_fullname}
    \bibliography{main}
}

\end{document}

\maketitle
\thispagestyle{empty} 
\renewcommand{\thefootnote}{\fnsymbol{footnote}}
\footnotetext[1]{: Equal contributions}
\renewcommand{\thefootnote}{\faEnvelope[regular]}
\footnotetext[2]{: Corresponding authors}
\section{Changes in Layer Precision During Training}
%%%%%%%%%%%%%%%%%%%%%%%%%%%%%%%%%%%%%%%%%%%%%%%%%%%%%%%%%%%%%%%%%%%%
\begin{figure}[b]
    \centering
    \begin{subfigure}{1\linewidth}
         \includegraphics[width=\linewidth]{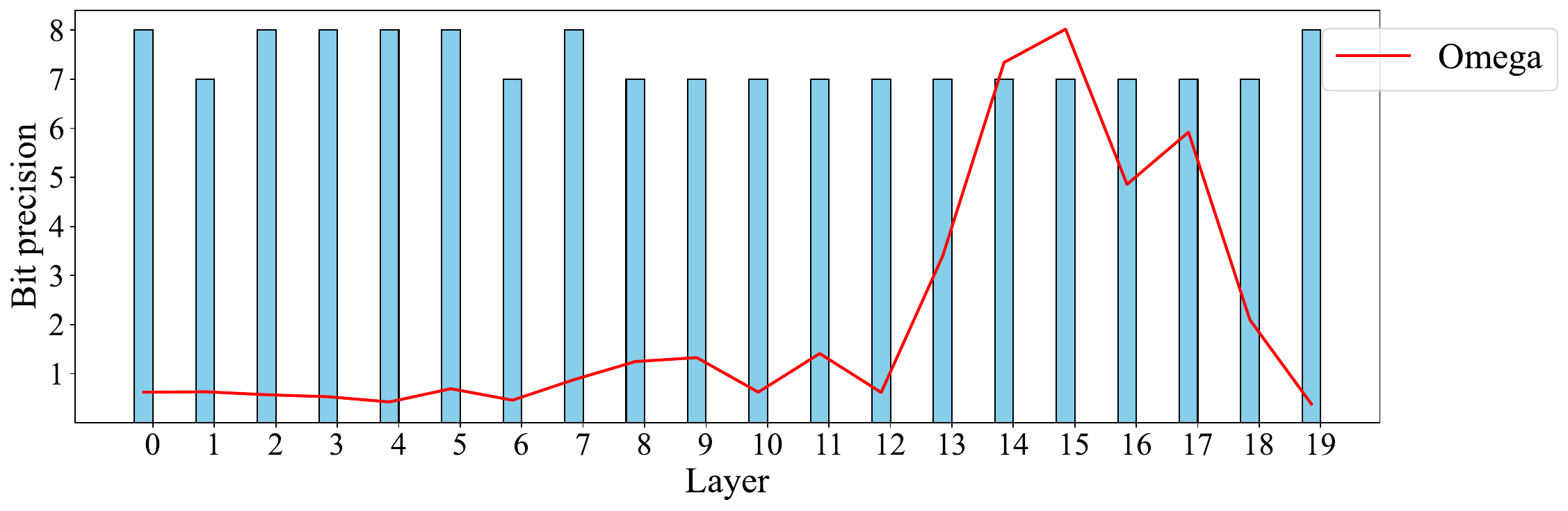}
        \caption{First Pruning Step (Compression Rate: 4.54x)}
        \label{fig:bit_list1}
    \end{subfigure} 
    \begin{subfigure}{1\linewidth}
        \includegraphics[width=\linewidth]{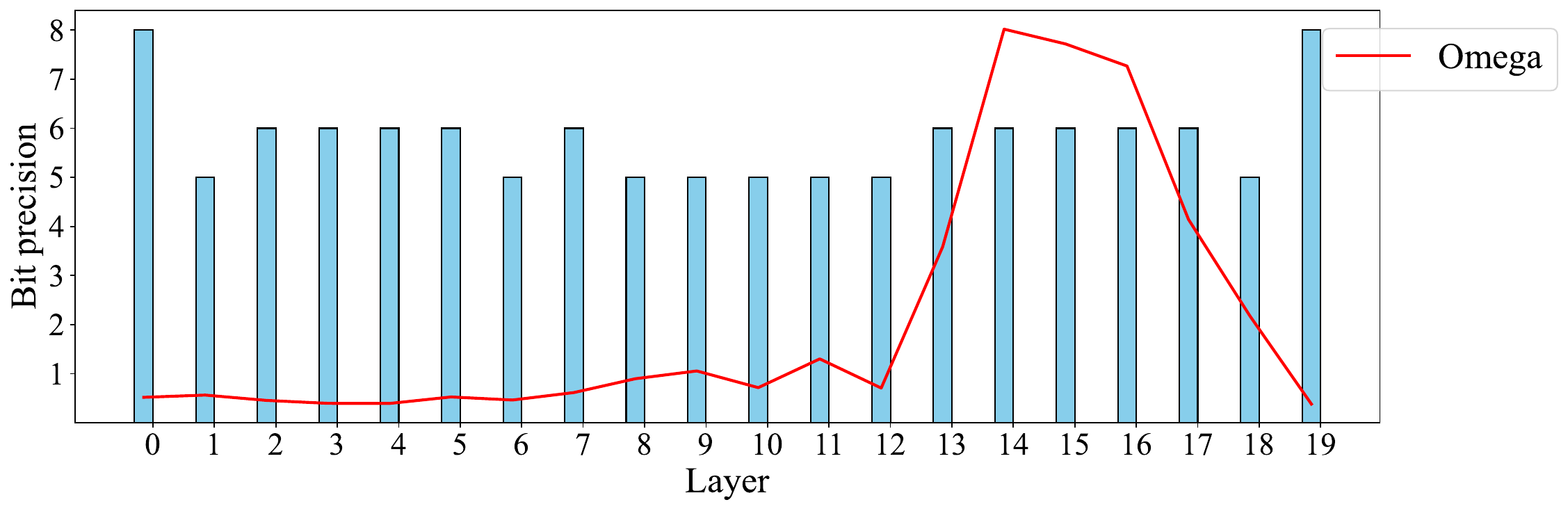}
        \caption{ Second Pruning Step (Compression Rate: 5.63x)}
        \label{fig:bit_list2}
    \end{subfigure}
    \begin{subfigure}{1\linewidth}
        \includegraphics[width=\linewidth]{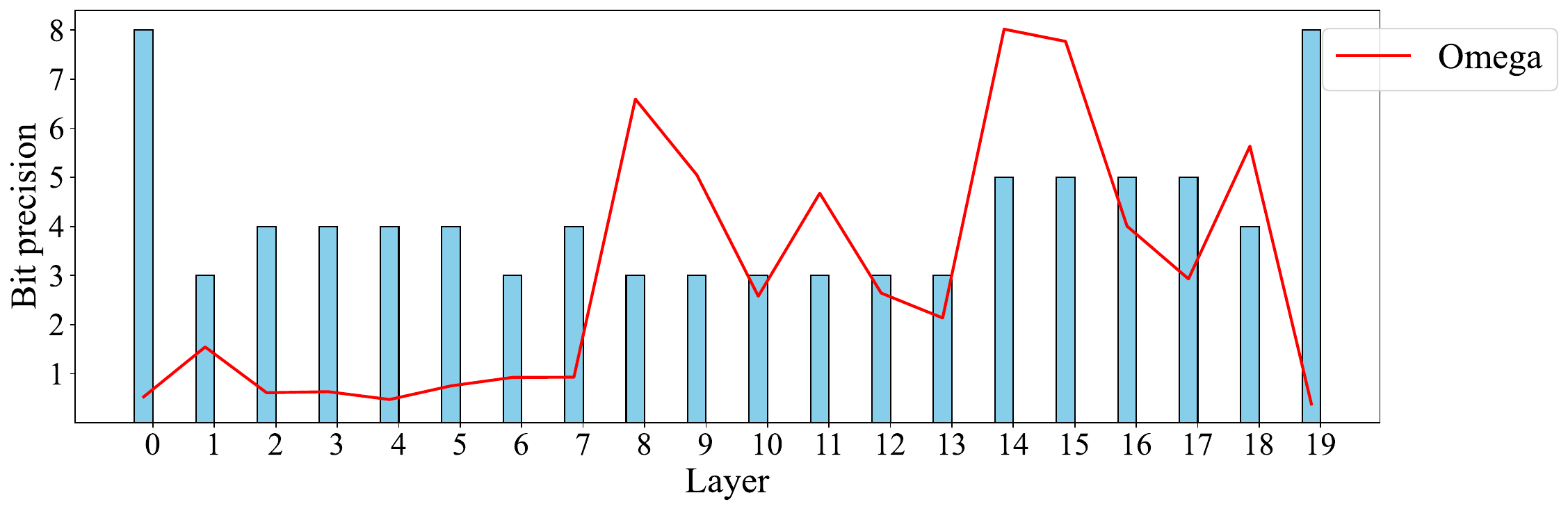}
        \caption{Third Pruning Step (Compression Rate: 7.20x)}
        \label{fig:bit_list3}
    \end{subfigure}
\end{figure}
%%%%%%%%%%%%%%%%%%%%%%%%%%%%%%%%%%%%%%%%%%%%%%%%%%%%%%%%%%%%%%%%%%%%
\begin{figure}[t]
\ContinuedFloat
    \begin{subfigure}{1\linewidth}
        \includegraphics[width=\linewidth]{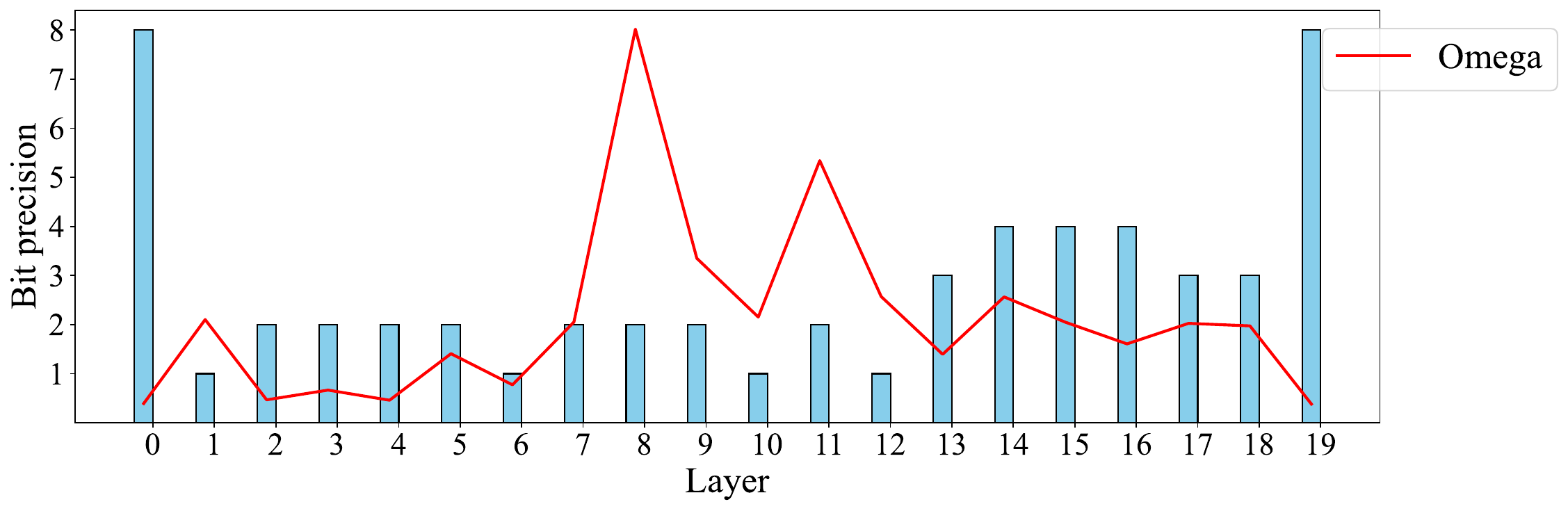}
        \caption{Fourth Pruning Step (Compression Rate: 10.30x)}
        \label{fig:bit_list4}
    \end{subfigure}
    \begin{subfigure}{1\linewidth}
        \includegraphics[width=\linewidth]{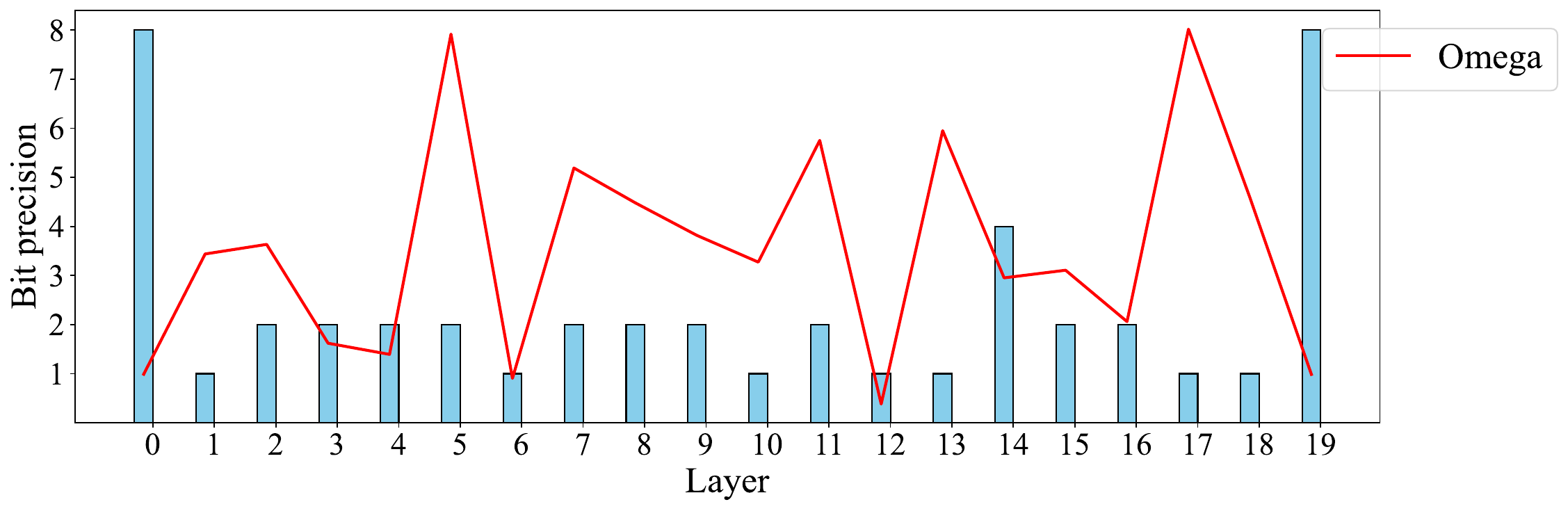}
        \caption{Last Pruning Step (Compression Rate: 17.11x)}
        \label{fig:bit_list5}
    \end{subfigure}
    \caption{Changes in Layer Precision and Omega During Training on ResNet-20}
    \label{fig:bit_list}
\end{figure}
%%%%%%%%%%%%%%%%%%%%%%%%%%%%%%%%%%%%%%%%%%%%%%%%%%%%%%%%%%%%%%%%%%%%
Fig.~\ref{fig:bit_list} illustrates how Omega values and bit precision change across layers during the training process of ResNet-20. Our Hessian Aware Aggressive Pruning method dynamically assigns prune bits as either 1-bit or 2-bit based on Omega values, enabling efficient bit reduction while maintaining model performance.

In the first pruning step Fig.~\ref{fig:bit_list1}, Hessian Aware Aggressive Pruning is not yet applied, so all layers retain a prune bit of 1. After pruning occurs, the Hessian is calculated, and prune bits are dynamically reassigned as either 1-bit or 2-bit depending on Omega values.

As training progresses to the second pruning step (Fig.~\ref{fig:bit_list2}), pruning occurs across most layers. However, as observed in Fig.~\ref{fig:bit_list1} and Fig.~\ref{fig:bit_list2}, layers that were assigned a prune bit of 2 in the first step undergo a rapid reduction in bit precision. For example, Layer index 11 is reduced from 7-bit to 5-bit, whereas Layer index 13 decreases from 7-bit to 6-bit.

This iterative pruning step continues, progressively refining bit precision across layers. Ultimately, the model achieves a compression rate of 17.11×, as depicted in Fig.~\ref{fig:bit_list5}.

\section{Final Bit Schemes}
\begin{figure}[t]
    \centering
        \begin{subfigure}{0.43\textwidth}
        \centering
        \includegraphics[width=\linewidth]{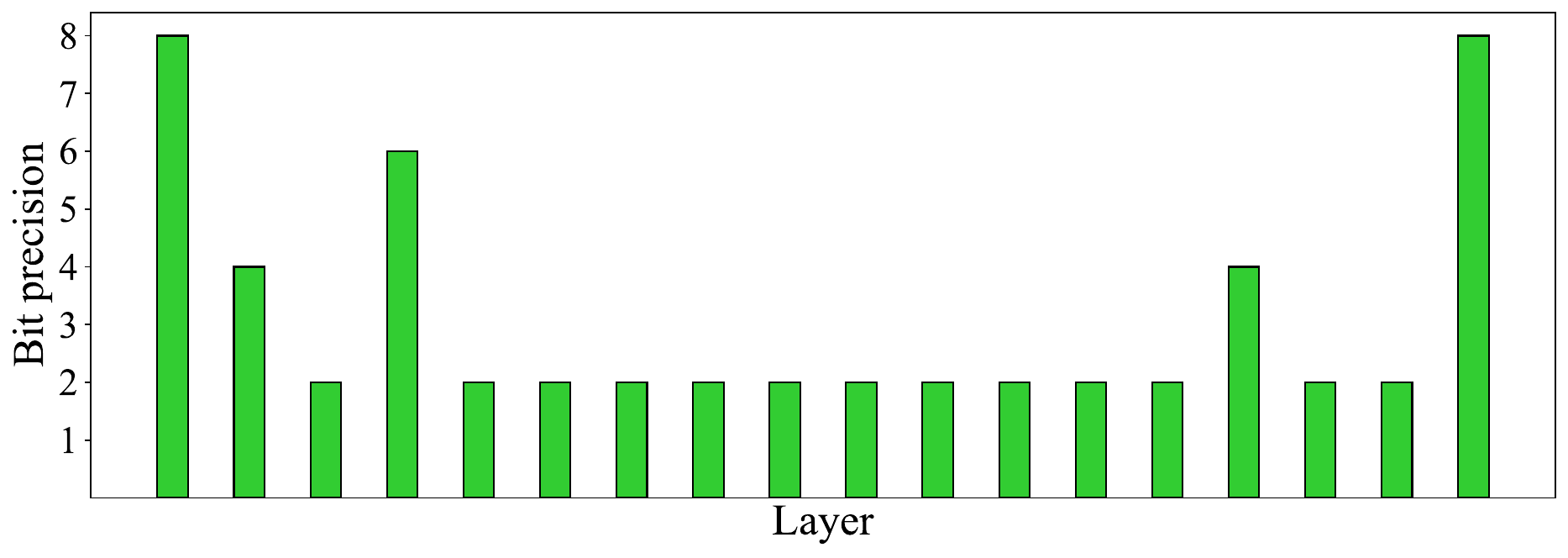}
        \caption{ResNet-18}
        \label{fig:bit_resnet18}
    \end{subfigure}
    \begin{subfigure}{0.43\textwidth}
        \centering
        \includegraphics[width=\linewidth]{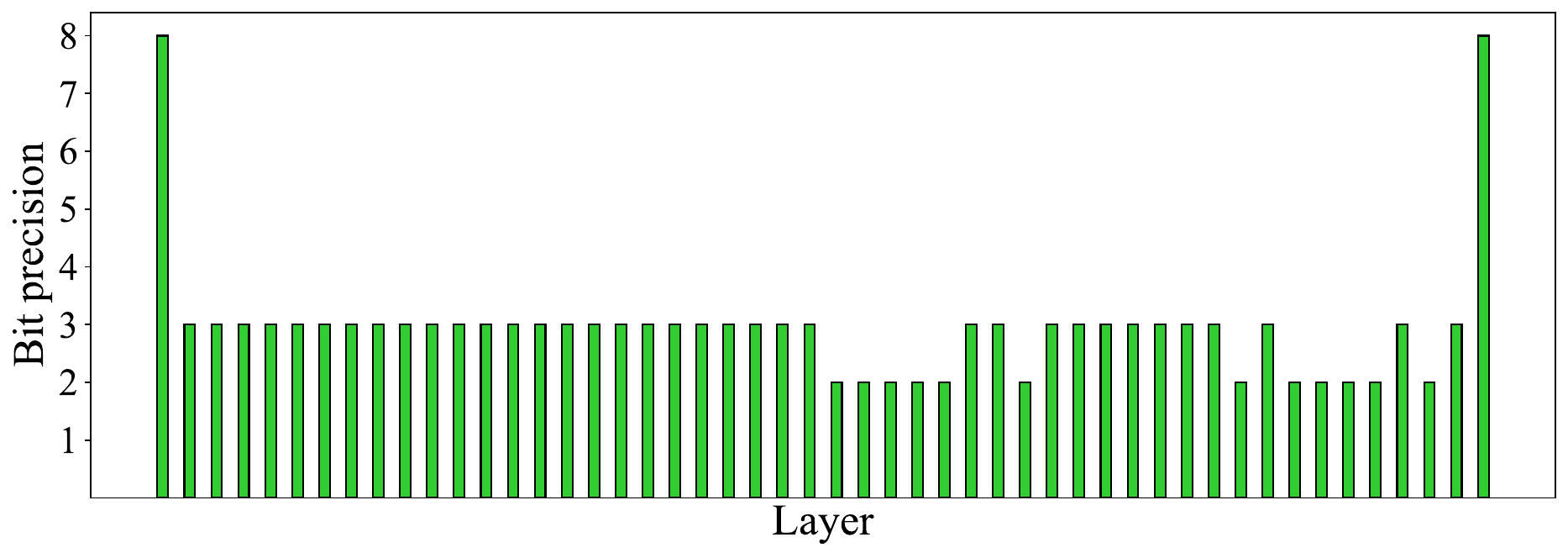}
        \caption{ResNet-50}
        \label{fig:bit_resnet50}
    \end{subfigure}
\caption{Final bit schemes of ResNet-18 and ResNet-50 after 100 epochs of training with MSQ.}

        \label{fig:resnet-bit-scheme}
\end{figure}

\begin{figure}[t]

    \begin{subfigure}{0.48\textwidth}
        \centering
        \includegraphics[width=\linewidth]{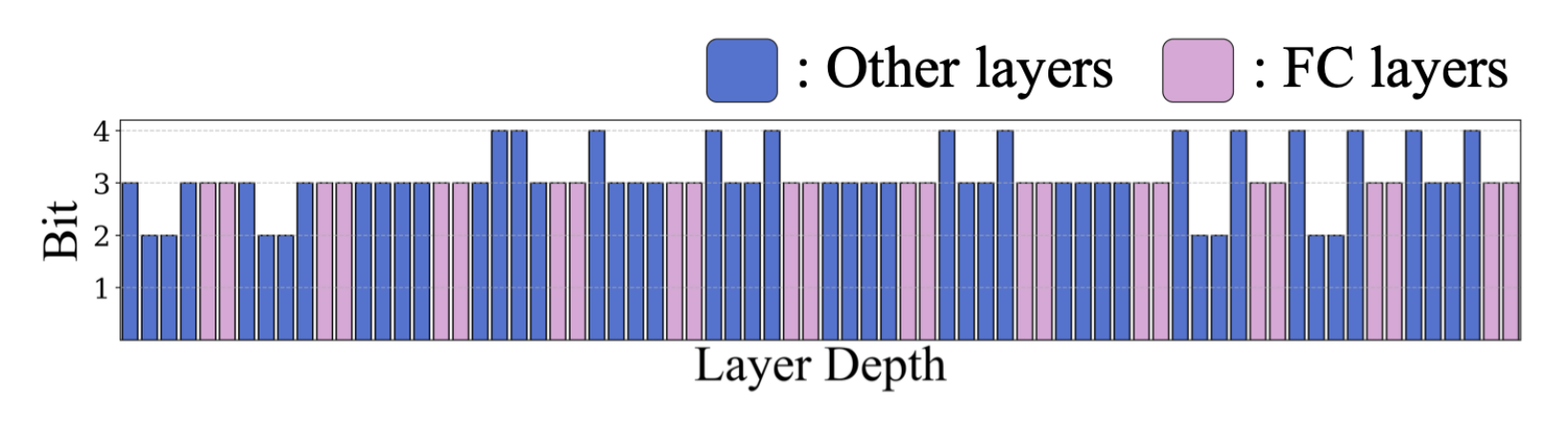}
        \caption{DeiT-T}
        \label{fig:bit_deit-t}
    \end{subfigure}
    \hfill
    \begin{subfigure}{0.48\textwidth}
        \centering
        \includegraphics[width=\linewidth]{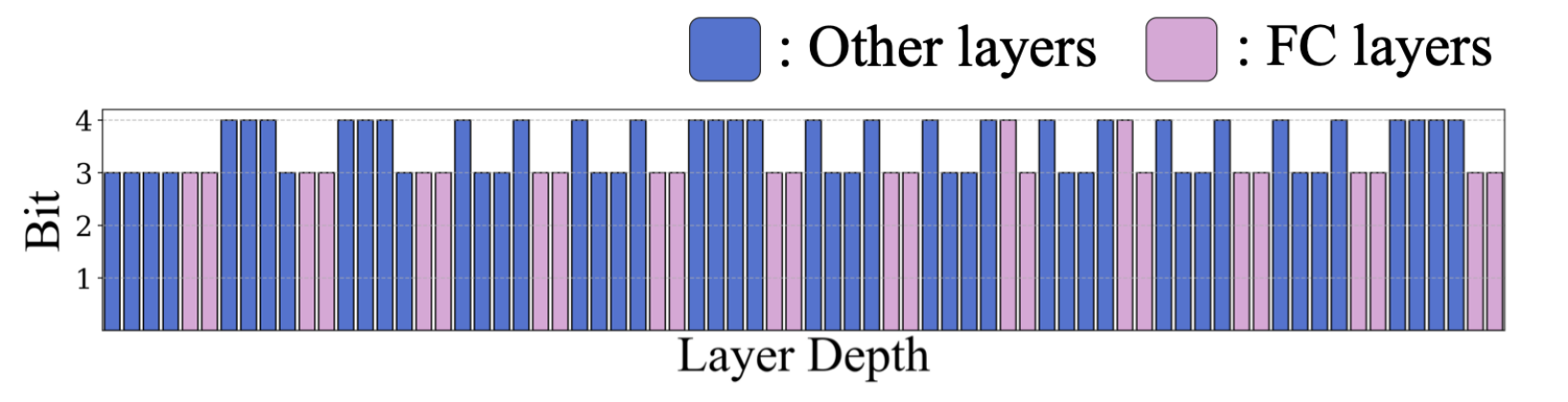}
        \caption{DeiT-S}
        \label{fig:bit_deit-s}
    \end{subfigure}
    \caption{Final bit schemes of DeiT-T and DeiT-S after fine-tuning with MSQ. The FC layers are colored in pink.}
    \label{fig:vit-bit-scheme}
\end{figure}

The final bit schemes for ResNet-18, ResNet-50, and ViT models, which were not included in the main text, are provided in this section.
Fig.~\ref{fig:resnet-bit-scheme} presents the bit schemes for ResNet-18 and ResNet-50. Specifically, Fig.~\ref{fig:bit_resnet18} achieves a compression ratio of 11.84×, whereas Fig.~\ref{fig:bit_resnet50} achieves a compression ratio of 10.89×.
Fig.~\ref{fig:vit-bit-scheme} presents the bit schemes for DeiT-T and DeiT-S. Fig.~\ref{fig:bit_deit-t} achieves a compression ratio of 10.54×, whereas Fig.~\ref{fig:bit_deit-s} achieves a compression ratio of 9.58×. Notably, the fully connected (FC) layers, which are observed to be pruned more aggressively.

%%%%%%%%%%%%%%%%%%%%%%%%%%%%%%%%%%%%%%%%%%%%%%%%%%%%%%%%%%%%%%%%%%%%%%%%%%%
%%%%%%%%%%%%%%%%%%%%%%%%%%%%%%%%%%%%%%%%%%%%%%%%%%%%%%%%%%%%%%%%%%%%%%%%%%%

%%%%%%%%%%%%%%%%%%%%%%%%%%%%%%%%%%%%%%%%%%%%%%%%%%%%%%%%%%%%%%%%%%%%%%%%%%%

\section{Additional Experiments on ViT}
To further validate the scalability of our method on larger transformer-based models, we conduct supplementary experiments on ViT-Base-Patch16-224. While the main paper presents results on compact vision transformers such as DeiT-T, DeiT-S, and Swin-T, as shown in  evaluate on a substantially larger Table~\ref{table:vit_cifar100_quant} architecture with 86.6M parameters.

\begin{table}[tb]
\centering
\caption{\mbox{Evaluation on ViT-Base-Patch16-224 using CIFAR-100.}}
\label{table:vit_cifar100_quant}
\vspace{-0.5em}
\begin{minipage}[t]{0.63\columnwidth}
    \centering
    \scriptsize
    \begin{tabular}{l|c|c|c}
        \toprule
        Method & W-Bits & Comp(×) & Acc(\%) \\
        \midrule
        FP       & 32   & 1.00   & 92.06 \\
        \arrayrulecolor{gray}\hline\arrayrulecolor{black}
        DoReFa  & 4    & 8.00   & 90.20 \\
        \rowcolor{gray!20}
        \textbf{MSQ}      & MP   & \textbf{9.14}   & \textbf{91.45} \\
        \bottomrule
    \end{tabular}
\end{minipage}%
\hfill
\begin{minipage}[t]{0.33\columnwidth}
    \centering
    \scriptsize
    \begin{tabular}{cc}
        \toprule
        \multicolumn{2}{c}{Hyperparameters}  \\
        \midrule
        Epochs & 50 \\
        $\lambda$ & 5e\textminus5 \\
        $I$ & 5 \\
        $\alpha$ & 0.3 \\
        \bottomrule
    \end{tabular}
\end{minipage}
% \vspacebot
\vspace{-1.5em}
\end{table}
%%%%%%%%%%%%%%%%%%%%%%%%%%%%%%%%%%%%%%%%%%%%%%%%%%%%%%%%%%%%%%%%%%%%%%%%%%%
\section{Hyperparameter Details}

The main hyperparameters of MSQ include $\lambda$, which controls the L1 regularization strength, $\alpha$, the pruning threshold that determines pruning decisions based on each layer’s LSB non-zero rate, and $I$, the pruning interval. 
The pruning interval $I$ is crucial for guiding LSB sparsification and facilitating accuracy recovery after pruning. Our experimental settings can be found in Table~\ref{table:hyperparameter}.

\begin{figure}[b]
    \centering
    \includegraphics[width=0.8\linewidth]{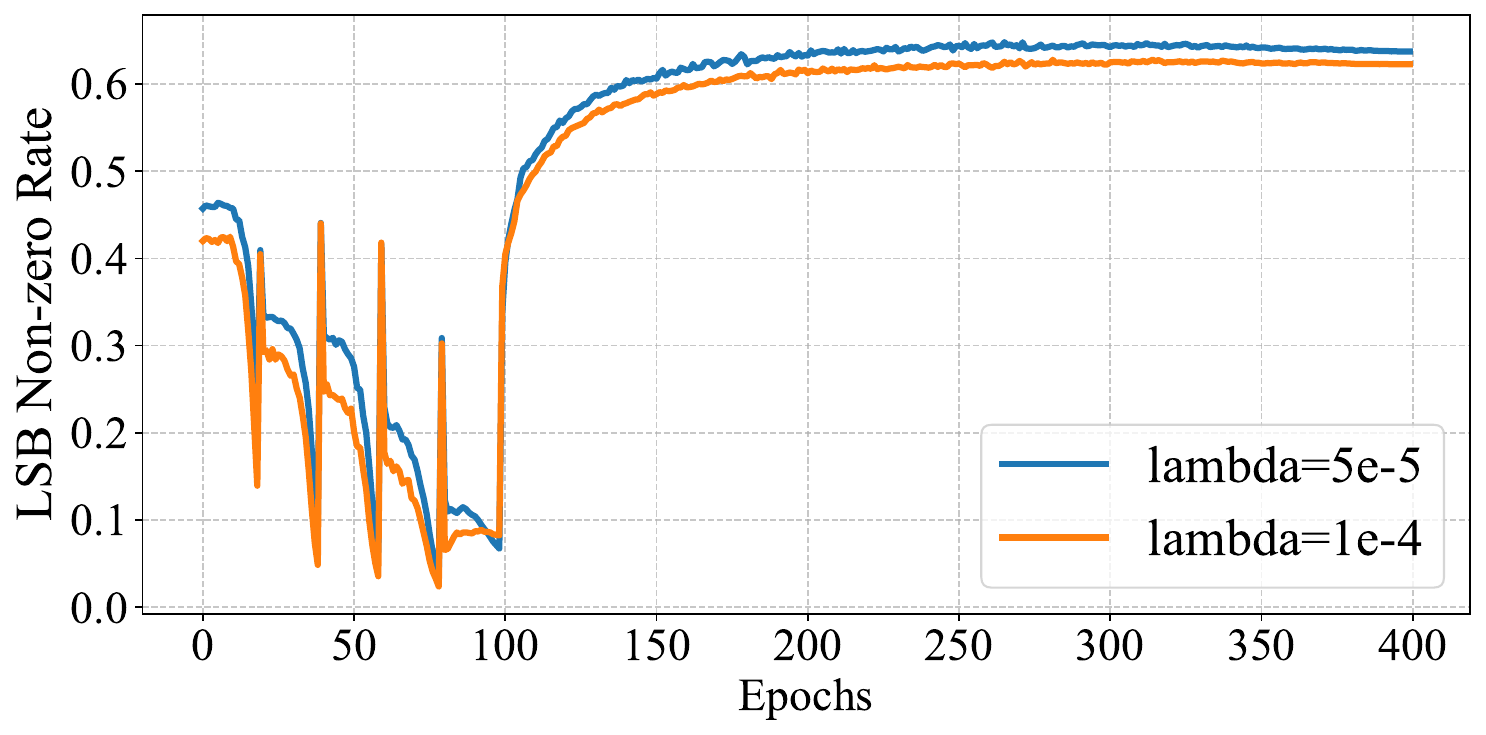}
    \caption{Comparison of \(\lambda = 5e{-5}\) and \(\lambda = 1e{-4}\). The LSB non-zero rate is relatively smaller when \(\lambda = 1e{-4}\).
}
    \label{fig:lambda}
\end{figure}

Reducing $\lambda$ decreases the regularization strength on LSBs, leading to less sparsity. Conversely, increasing $\lambda$ strengthens the regularization effect, deriving higher sparsity as depicted on Fig.~\ref{fig:lambda}. 
However, setting $\lambda$ too high may cause excessive LSB regularization, potentially degrading accuracy. Thus, it is essential to carefully tune $\lambda$ and the pruning threshold $\alpha$ to balance sparsity and accuracy effectively. 
%%%%%%%%%%%%%%%%%%%%%%%%%%%%%%%%%%%%%%%%%%%%%%%%%%%%%%%%%%%%%%%%%%%%%%%%%%%%
\begin{table}[b] 
     \centering
    \caption{Hyperparameter settings for pruning on our experiments. $\lambda$ represents the L1 regularization strength, $\alpha$ determines the pruning decision based on each layer's non-zero rate, and $I$ denotes the pruning interval.}
\begin{tabular}{l|ccc}
    \toprule
    \small
    Network             & $\lambda$ & $\alpha$  & $I$  \\
    \midrule
    ResNet-20           & 5e-5      & 0.3       & 20   \\
    ResNet-18           & 5e-5      & 0.3       & 10   \\
    ResNet-50           & 5e-5      & 0.3       & 10   \\
    DeiT-T              & 8e-6      & 0.35      & 5    \\
    DeiT-S              & 5e-6      & 0.35      & 8    \\
    Swin-T              & 5e-6      & 0.35      & 8    \\
    MobileNetV3-Large   & 5e-5      & 0.3       & 5  \\
    \bottomrule
\end{tabular}

     \label{table:hyperparameter}
 \end{table}
%%%%%%%%%%%%%%%%%%%%%%%%%%%%%%%%%%%%%%%%%%%%%%%%%%%%%%%%%%%%%%%%%%%%%%%%%%%

%%%%%%%%%%%%%%%%%%%%%%%%%%%%%%%%%%%%%%%%%%%%%%%%%%%%%%%%%%%%%%%%%%%%%%%%%%%